\documentclass[sigconf]{acmart}
\usepackage{algorithm}
\usepackage{algorithmic}
\usepackage{multirow}
\usepackage{amssymb}
\usepackage{subcaption}

\renewcommand\footnotetextcopyrightpermission[1]{} 

\AtBeginDocument{%
	\providecommand\BibTeX{{%
			\normalfont B\kern-0.5em{\scshape i\kern-0.25em b}\kern-0.8em\TeX}}}

\setcopyright{none}

\acmConference[None]{MM '20: ACM Multimedia}
\acmBooktitle{None}
\acmPrice{15.00}
\acmISBN{978-1-4503-XXXX-X/18/06}

\acmSubmissionID{390}

\begin{document}
	
	\title{Localizing Interpretable Multi-scale informative Patches Derived from Media Classification Task}
	
	\author{Chuanguang Yang}
	\email{yangchuanguang@ict.ac.cn}
	\affiliation{%
	  \institution{Institute of Computing Technology, Chinese Academy of Sciences}
	  \institution{University of Chinese Academy of Sciences}
	  \city{Beijing}
	  \country{China}
	}

	\author{Zhulin An}
	\authornote{Corresponding Author.}
	\email{anzhulin@ict.ac.cn}
	\affiliation{%
		\institution{Institute of Computing Technology, Chinese Academy of Sciences}
		\institution{University of Chinese Academy of Sciences}
		\city{Beijing}
		\country{China}
	}

	\author{Xiaolong Hu}
\email{huxiaolong18g@ict.ac.cn}
\affiliation{%
	\institution{Institute of Computing Technology, Chinese Academy of Sciences}
	\institution{University of Chinese Academy of Sciences}
	\city{Beijing}
	\country{China}
}

	\author{Hui Zhu}
\email{zhuhui@ict.ac.cn}
\affiliation{%
	\institution{Institute of Computing Technology, Chinese Academy of Sciences}
	\institution{University of Chinese Academy of Sciences}
	\city{Beijing}
	\country{China}
}

	\author{Yongjun Xu}
\email{xyj@ict.ac.cn}
\affiliation{%
	\institution{Institute of Computing Technology, Chinese Academy of Sciences}
	\institution{University of Chinese Academy of Sciences}
	\city{Beijing}
	\country{China}
}
	%
	%
	%
	%
	%
	%
	%
	
	\renewcommand{\shortauthors}{Anonymous authors.}
	
	\begin{abstract}
		Deep convolutional neural networks (CNN) always depend on wider receptive field (RF) and  more complex non-linearity to achieve state-of-the-art performance, while suffering the increased difficult to interpret how relevant patches contribute the final prediction. In this paper, we construct an interpretable AnchorNet equipped with our carefully designed RFs and linearly spatial aggregation to provide patch-wise interpretability of the input media meanwhile localizing multi-scale informative patches only supervised on media-level labels without any extra bounding box annotations. Visualization of localized informative image and text patches show the superior multi-scale localization capability of AnchorNet. We further use localized patches for downstream classification tasks across widely applied networks. Experimental results demonstrate that replacing the original inputs with their patches for classification can get a clear inference acceleration with only tiny performance degradation, which proves that localized patches can indeed retain the most semantics and evidences of the original inputs.
	\end{abstract}

	
	
	
	\maketitle
	
	\section{Introduction}
	Although deep convolutional neural networks (CNN) achieve superior performance across a broad range of  tasks, such as image classification \cite{he2016deep,huang2017densely}, text classification \cite{DBLP:conf/emnlp/Kim14} and other multi-media applications \cite{DBLP:conf/mm/WangX0HSS19,DBLP:conf/mm/WangSWS19}. However, the decision of the CNN for a given media still suffers severe difficulty to interpret, thus challenging some critical applications, such as healthcare, automatic driving and criminal justice \cite{rudin2019stop}. Why the modern CNNs are difficult to interpret? On the one hand, the design of CNN is always equipped with large accumulated receptive field (RF) along with multiple paddings throughout the CNN, so the evidence of location in the high-level feature map is unable to determine the scanned region of the input media, e.g. the RF of ResNet-50 is $443\times 443$, which is much larger than the size of input image ($224\times 224$) due to the intermediate paddings. On the other hand, the classifier fully-connected (FC) layer further builds complex dependencies among hidden activations and various patches of the input image, making the impossible interpretation of the contribution of each patch. 
	
	\begin{figure}[tbp]  
		\centering  
		\includegraphics[width=1\linewidth]{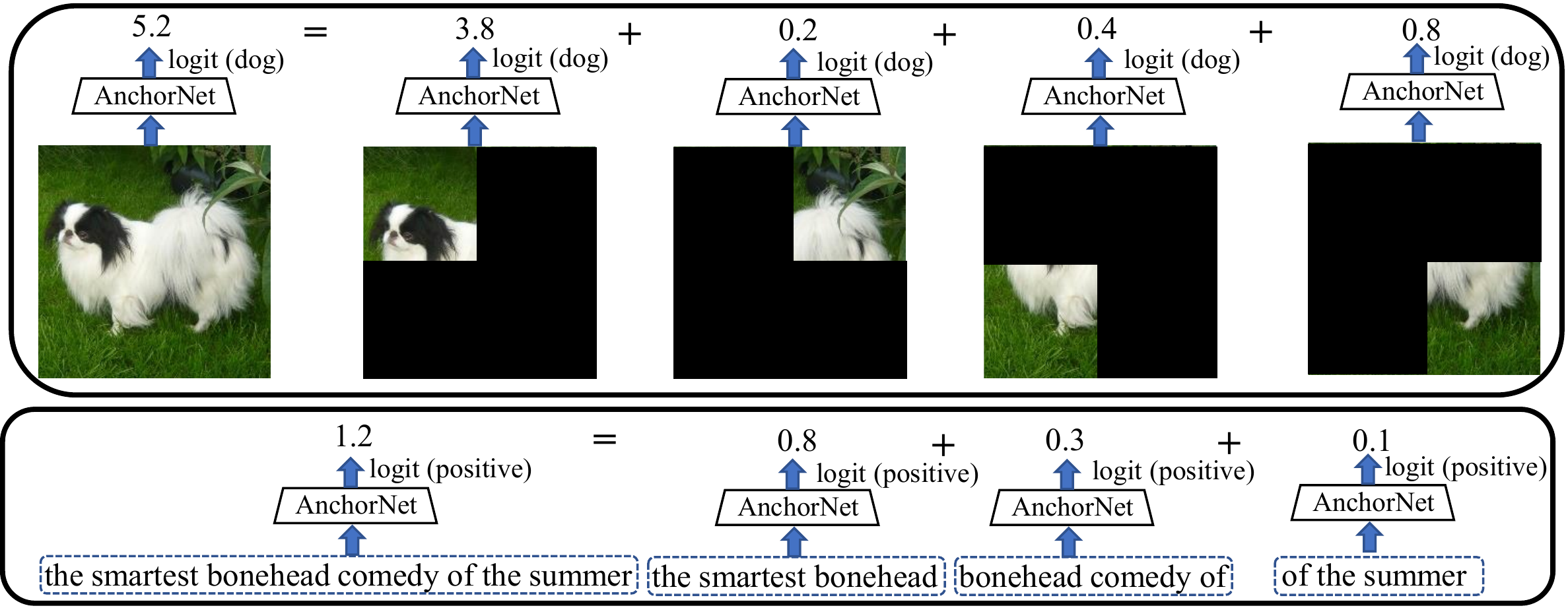}
		\caption{Illustrations of disentangled patch-wise contributions for final decision towards image (\emph{top}) and text (\emph{bottom}). }  
		\label{independ}
	\end{figure}
	
	To address these dilemmas, we design an interpretable CNN framework called AnchorNet attempting to estimate how patches in the input media contribute the final decision. On the one hand, we carefully design the RF without any paddings for each convolutional layer such that the accumulated RF (i.e. patch size) is smaller than the size of input media. On the other hand, we perform a spatially linear aggregation  (i.e., a simple average) without FC layer on the class-specific feature map before softmax layer. Benefit from the above techniques, each high-level spatial location can be mapped back to the corresponding patch within the input media, thus the contributions of patch-wise evidences can be transparently determined according to the activation of high-level locations. Meanwhile, the decision of each patch is thus disentangled and independent of the others, which we illustrate in Figure \ref{independ}.
	
	It is widely known that real objects usually have various scales along with coarse- or fine-grained features, so AnchorNet are equipped with three localization branches with various accumulated RFs so as to adaptively capture multi-scale informative patches. We instantiate AnchorNet-I and AnchorNet-T for image and text localization, respectively. Visualization of localized informative image patches across the ImageNet dataset \cite{deng2009imagenet} can be on a par with the localization performance of object detection methods, while our localized process is more efficient due to the single supervision on classification loss. Localized text patches can also capture the crucial evidences for sentiment recognition on MR \cite{DBLP:conf/acl/PangL04} dataset. We further observe that an informative patch in the image may provide confused evidences that not contribute to the prediction of its correct class, which interprets the cause of misclassification. In addition, we provide the reasonable interpretation for the case of misclassification between different localization branches in terms of RF and feature extraction. 
	
	To demonstrate that localized patches can retain the effective semantics of the original media, we perform downstream image and text classification tasks based on localized patches using widely applied networks. The performances of downstream classifiers only have a tiny drop while obtaining a clear acceleration, which indicates that the localization capability of AnchorNet is indeed remarkable and may bring potential benefits of speeding up inference. For the reason that the spatial arrangement of patches do not affect the final decision, we thus prove that the performance of localization and classification is robust to the texturised images \cite{DBLP:journals/corr/GatysEB15a}, while humans always suffer great difficulty to recognize them.

	In brief, our contributions mainly lie in four folds:
	\begin{itemize}
		\item We construct an interpretable AnchorNet to adaptively localize multi-scale informative patches by carefully designed RFs and linear aggregation for the given media.
		\item Localization is efficient due to the single supervision on media-level labels without any bounding box annotations.
		\item We provide the reasonable interpretation of misclassification in terms of patch-wise evidences.
		\item The pipeline of designing upstream AnchorNet for downstream model may inspire the patch-wise classification to the field of network acceleration in future.
	\end{itemize}
	
	\section{Related Work}
	\textbf{Patch-based deep networks.} Before the emergence of deep CNN, bag-of-feature (BoF) models dominantly perform on recognition tasks by providing a set of local features. After combined with CNN, patch-based deep features have been researched extensively in object classification \cite{DBLP:journals/pami/WeiXLHNDZY16,DBLP:journals/pr/TangWHBL17,DBLP:journals/corr/TangWSBLT16}, scene recognition \cite{DBLP:conf/cvpr/ArandjelovicGTP16, DBLP:journals/cin/FengLW17} and image retrieval \cite{DBLP:conf/mm/CaoHS17,DBLP:conf/cvpr/NgYD15}. However, the above works do not carefully calculate the RF to guarantee the strict mapping from each high-level location to the corresponding patch in the input image. They only perform vanilla mapping by relative proportion of the spatial size between high-level feature map and input image, so they may ignore that each high-level location has seen much larger region than their mapping region, then leading to be uninterpretable for the contribution of each patch.
	
	\textbf{Interpretable CNNs.} Some previous works perform pixel-wise interpretability on decision-making \cite{DBLP:conf/kdd/Ribeiro0G16,DBLP:conf/iccv/FongV17,selvaraju2017grad}. In this vein, Zhang \emph{et al.} \shortcite{DBLP:conf/cvpr/ZhangWZ18a} construct an interpretable CNN to explain its logic at the object-part
	level, but not for patch-level. Our work relates closely to ProtoPNet \cite{DBLP:conf/nips/ChenLTBRS19} and Saccader \cite{elsayed2019saccader}, both of them aim to study  patch-wise interpretability. ProtoPNet  \cite{DBLP:conf/nips/ChenLTBRS19} interprets prototypical patches by applying L2 distance between latent representations and prototypes, but Chen \emph{et al.} \shortcite{DBLP:conf/nips/ChenLTBRS19} neglect that similar latent representations still exist some gaps of human-interpretable features. Moreover, ProtoPNet is only suitable for fine-grained recognition, so it may be not general in practice. Saccader ~\cite{elsayed2019saccader} introduces a hard attention module to estimate the relevance of various image patches, but suffers an optimization difficulty for hard attention. Moreover, the complexity of Saccader is large along with policy gradient optimization, leading to the heavy computational costs. Our AnchorNet solves the above regrets that provides human-interpretable features by linearly feature aggregation as well as mapping, and can apply to both coarse- and fine-grained recognition. Moreover, AnchorNet is quite light-weight and can be easy to optimized by standard stochastic gradient descent (SGD) method via a end-to-end manner.

	\textbf{Localizing informative features.} Some previous works perform interpretability analysis of CNN by visualizing the semantic feature heatmap, mainly divided into response-based \cite{zhou2016learning, fukui2019attention} or gradient-based \cite{springenberg2014striving,selvaraju2017grad,smilkov2017smoothgrad} manners. However, visual explanation only displays the semantic region that is unable to explain the reasoning process and implement  the downstream classification task due to its irregular shape. Some seemingly similar but essentially different approaches are region proposal models for object detection ~\cite{girshick2015fast,ren2015faster,he2017mask}, which typically aggregate the information far beyond the region proposals, and use ground-truth bounding boxes for training. Unlike these works, our AnchorNet is only supervised by image-level labels and extracts local features in the fixed regions that are strictly spatial alignment to the initial input image. Zhou \emph{et al.}~\shortcite{zhou2016learning} implement object localization without supervision on any bounding box annotations, which shares the similarity to us of training by image-level labels. However, the information is still gathered from the whole image instead of strictly defined patches, hence the contributions of various patches to final prediction would get tangled. Additionally, AnchorNet utilizes one or more patches with the same size to cover the object, which are quite useful for downstream model meanwhile obtaining a good performance. 
	
	\textbf{Attention mechanisms.} Channel attention \cite{xhu2018squeeze,DBLP:conf/cvpr/LiW0019} are widely used in recognition tasks for capturing channel-wise dependencies so as to improve the performance of CNN. Beyond channel, spatial attention \cite{woo2018cbam,fukui2019attention,elsayed2019saccader} is usually introduced to highlight the spatially semantic locations . AnchorNet differs from those prior practices in that we apply multi-branch attention mechanism to perform multi-RF semantic localization and improve the capability of multi-scale feature representations. 
	
	\textbf{Inference acceleration.} Modern inference acceleration methods concentrate on model-based processing, which is mainly divided into two aspects: static pruning \cite{Hao2017Pruning,he2017channel,liu2019metapruning}, which aims to remove redundant structure of the model, or dynamic inference \cite{Gao2017Mult,wu2018blockdrop,veit2018convolutional}, which aims to only use a part of structure of the model conditioned on the input image. In this work, we localize informative local image patches over the whole image guided by light-weight AnchorNet, then the downstream heavy networks only need to process semantic feature patches, the total area of which is much smaller than the original image, thus producing a clear acceleration. Moreover, data-based localization is model-agnostic and thus can be regarded as the orthogonal and complementary .
	
	\section{Methodology}
	\subsection{Review of Feature Mapping}
	\label{Review}
	\begin{figure}[tbp]  
		\centering  
		\includegraphics[width=0.75\linewidth]{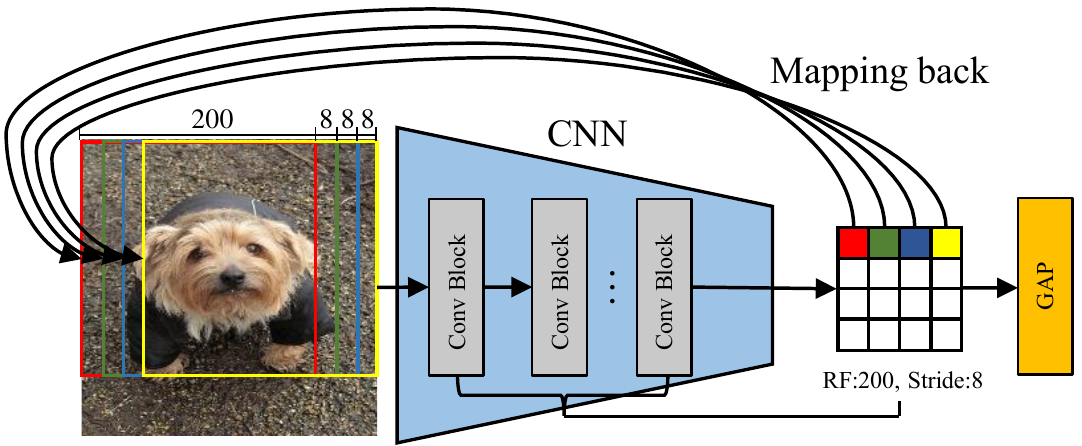}
		\caption{Example of patches mapping. We only depict the spatial dimension while omitting channels for better understanding of spatial mapping rule. Best viewed in color.}  
		\label{map}
	\end{figure}
	\begin{figure*}[tbp]  
		\centering  
		\includegraphics[width=0.95\linewidth]{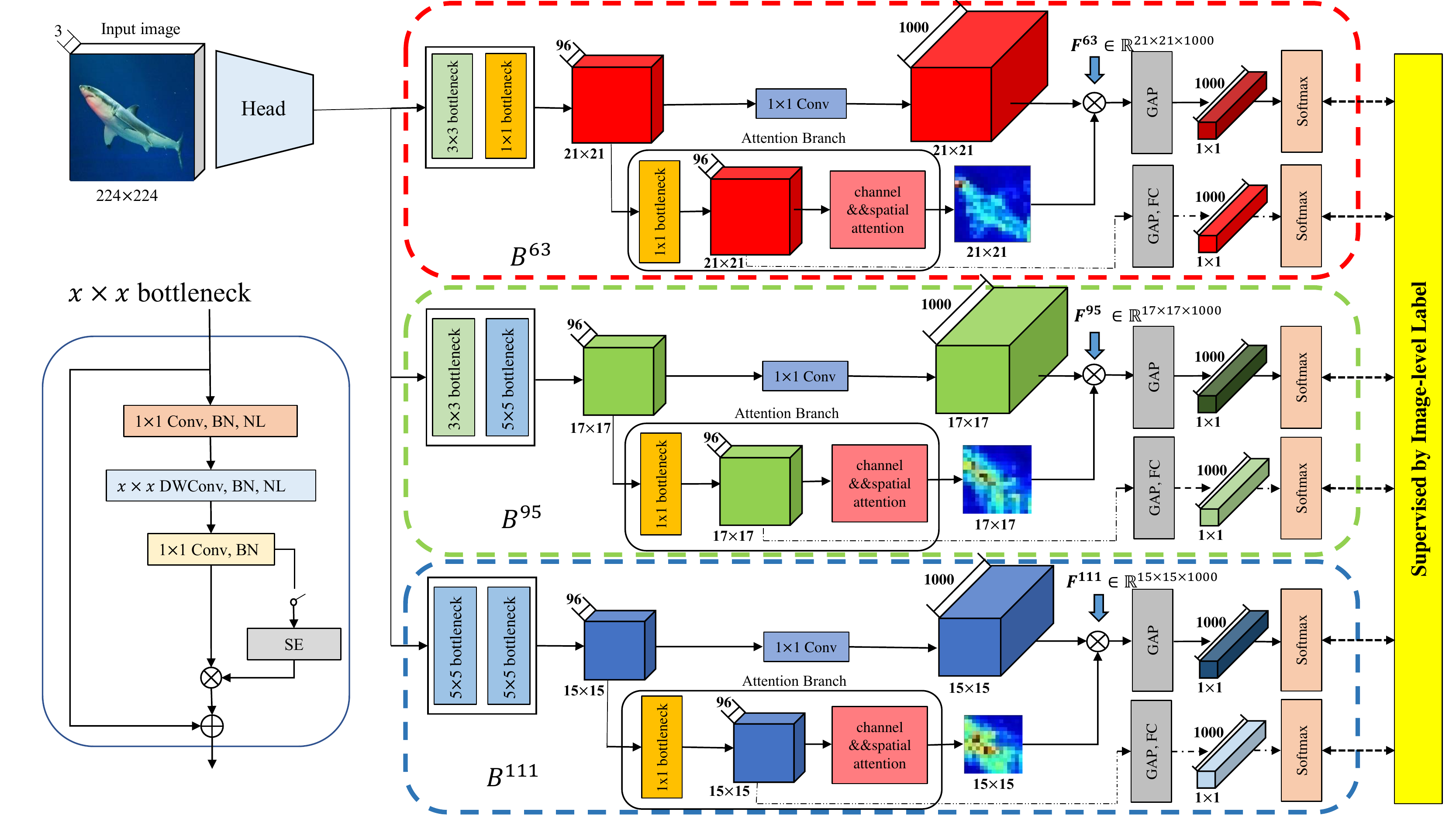}
		\caption{Illustration of the overall architecture of AnchorNet-I for 1000-classification on ImageNet. Three branches associated with various RFs of $63\times 63$, $95\times 95$ and $111\times 111$ after the shared head perform multi-scale patch localization, which we name them as $B^{63}$, $B^{95}$ and $B^{111}$, respectively. Sizes of spatial and channel are tagged below and over each 3D feature map, respectively.}  
		\label{anchornet}
	\end{figure*}
	Modern CNNs gradually decrease the spatial resolution for the input image by several convolutional blocks until the global average pooling (GAP) layer. Many hyperparameters in convolutional layer settings, e.g., kernel size, padding or stride can affect resolution size of the output. We set padding 0 across all the convolutional layers in AnchorNet, so each final spatial location of the feature map before GAP layer can be mapped to the input image exactly without the cases of beyond bounds. Assumed that an interpretable CNN model receives an image with $H\times W$ pixels as the input, and has accumulated $k\times k$ RF size and $s$ strides before GAP, we will obtain $[\left \lfloor (H-k)/s \right \rfloor+1] \times [\left \lfloor (W-k)/s \right \rfloor+1]$ spatial locations, where each location can be mapped back to a region with the size of $k\times k$, e.g. for an input image with $224\times 224$ pixels to the CNN model, which has accumulated $200\times 200$ RF size and $8$ strides before GAP, can generate a feature map with $4\times4=16$ spatial locations, as shown in Figure \ref{map}.

	\begin{table}[t]
		\centering
		\caption{The structural settings of head.}
		\begin{tabular}{c|c|c|c|c|c|c|c}  
			\toprule
			IR & Operator & Exp &Out &SE &NL&$s$&RF \\
			\midrule
			$224^{2}$       & conv2d,3$\times$3   &-  &16&-&HS&2&$3^{2}$   \\
			$111^{2}$      & bneck,3$\times$3& 16&16&-&RE&2&$7^{2}$   \\
			$55^{2}$      & bneck,3$\times$3& 72 &24&-&RE&2&$15^{2}$   \\
			$27^{2}$       & bneck,1$\times$1& 88 &24&-&RE&1&$15^{2}$   \\
			$27^{2}$       & bneck,1$\times$1& 96&40&\checkmark&HS&1&$15^{2}$   \\
			$27^{2}$      & bneck,1$\times$1& 240 &40&\checkmark&HS&1&$15^{2}$   \\
			$27^{2}$       & bneck,1$\times$1& 240 &40&\checkmark&HS&1&$15^{2}$   \\
			$27^{2}$       & bneck,1$\times$1& 120 &48&\checkmark&HS&1&$15^{2}$   \\
			$27^{2}$       & bneck,3$\times$3& 144 &48&\checkmark&HS&1&$31^{2}$   \\
			$25^{2}$       & bneck,3$\times$3& 288 &96&\checkmark&HS&1&$47^{2}$   \\
			\bottomrule
		\end{tabular}
		\label{head}
	\end{table}
	
	\begin{table}[t]
		\centering
		\caption{The structural settings of localization branches.}
		\begin{tabular}{c|c|c|c|c|c|c|c}  
			\toprule
			IR & Operator & Exp &Out &SE &NL&$s$&RF \\
			\midrule
			$23^{2}$& bneck,3$\times $3& 480&96&\checkmark&HS&1&$63^{2}$   \\
			$21^{2}$     & bneck,1$\times$1& 576&96&\checkmark&HS&1&$63^{2}$   \\
			$21^{2}$     & bneck,1$\times$1& 192&96&\checkmark&HS&1&$63^{2}$   \\
			\midrule
			$23^{2}$& bneck,3$\times$3& 480&96&\checkmark&HS&1&$63^{2}$   \\
			$21^{2}$     & bneck,5$\times$5& 576&96&\checkmark&HS&1&$95^{2}$   \\
			$17^{2}$     & bneck,1$\times$1& 192&96&\checkmark&HS&1&$95^{2}$   \\
			\midrule
			$23^{2}$& bneck,5$\times$5& 480&96&\checkmark&HS&1&$79^{2}$   \\
			$19^{2}$     & bneck,5$\times$5& 576&96&\checkmark&HS&1&$111^{2}$   \\
			$15^{2}$     & bneck,1$\times$1& 192&96&\checkmark&HS&1&$111^{2}$   \\
			\bottomrule
		\end{tabular}
		\label{tails}
	\end{table}
	\begin{figure}[tbp]  
		\centering  
		\includegraphics[width=1.0\linewidth]{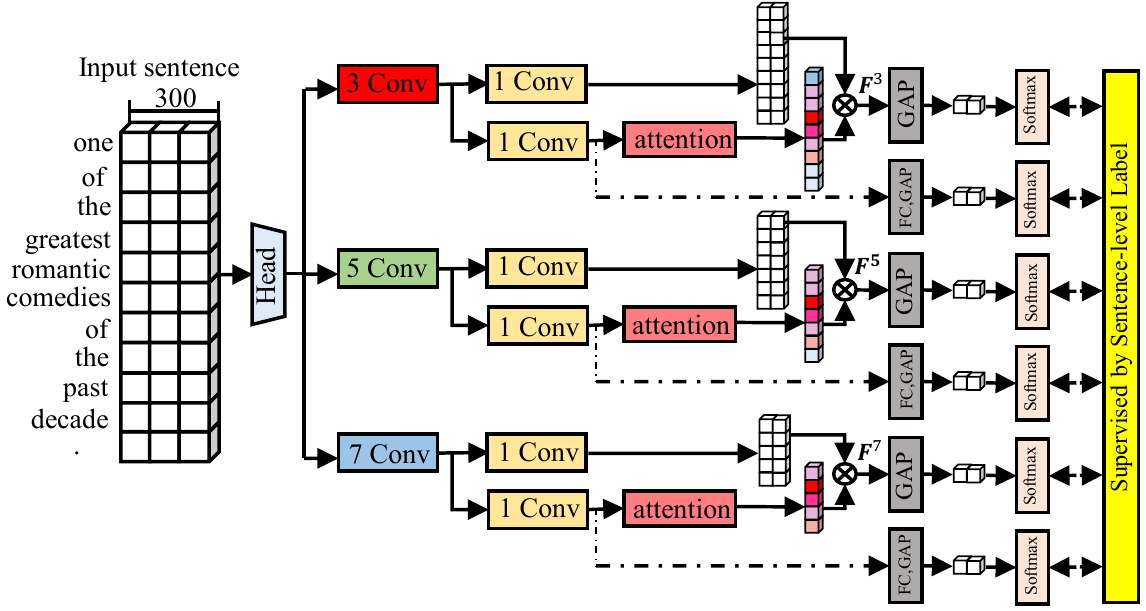}
		\caption{Illustration of the overall architecture of AnchorNet-T for 2-classification on MR. The structural details are shown in supplementary material section A.}  
		\label{AnchorNet-T}
	\end{figure}
	\subsection{AnchorNet}
	
	We develop an interpretable CNN framework called AnchorNet which can provide the contribution of each patch while adaptively localizing multi-scale informative patches conditioned on the input media. Specifically, we instantiate AnchorNet-I and AnchorNet-T for image and text localization respectively, as shown in Figure \ref{anchornet} and \ref{AnchorNet-T}, both of them are derived from classification, and contain the following components. Note that the main difference between AnchorNet-I and AnchorNet-T is the dimension of convolutions, because the spatial size of image feature is 2D, while that of text feature is 1D.

	\subsubsection{Head} For AnchorNet-I, the input image is firstly processed by a head to extract low-level features, the structure details of head is shown in Table \ref{head}, which is composed of several bottleneck units (bneck) \cite{howard2019searching}, where IR denotes the input resolution, Exp and Out denote the expansion and output channels, SE denotes whether there exists a SE block \cite{hu2018squeeze}, NL denotes the non-linearity, including $\rm h$-$\rm swish$ (HS)  \cite{howard2019searching} or $\rm ReLU$ (RE), $s$ denotes the stride of current convolution, RF denotes the accumulated RF size until the current layer. And we replace most $3\times 3$ convolutions with $1\times 1$ convolutions without any paddings so as to restrict accumulated RF throughout the head and thus guaranteeing the exactly patch mappings, the property of which is the most difference compared with popular networks. It is noteworthy that we perform less down-sampling compared with popular setting on ImageNet dataset  \cite{deng2009imagenet}, which can retain the higher spatial resolution of feature map so as to generate more patch mappings to the input image. For AnchorNet-T, head contains two convolutions to sequentially squeeze the channel dimension from 300 to 64 without changing the spatial size.

	\subsubsection{Localization Branch} AnchorNet constructs three branches to adaptively localize multi-scale informative patches along the spatial dimension after the shared head. To this end, bottlenecks with various kernel sizes are intentionally equipped to adjust the accumulated RF sizes of these branches individually in AnchorNet-I. Table \ref{tails} elaborates the information of accumulated RF, where the blocks sequentially correspond the localization branches $B^{63}$, $B^{95}$ and $B^{111}$ in Figure \ref{anchornet}, respectively. And the final row of each block denotes the bottleneck in attention branch. It means that three feature maps generated by the three localization branches would obtain a pixel-wise mapping patch size of $63\times 63$, $95\times 95$ or $111\times 111$ to the input image, respectively. Due to the accumulated stride of all the three branches is $2^{3}=8$, they can map to  $21^{2}=441$, $17^{2}=289$, $15^{2}=225$ possible informative patches for the original image with the size of $224\times 224$, respectively. The principle of mapping has been reviewed in section \ref{Review}. Similarly, AnchorNet-T is equipped with the accumulated RFs of 3, 5, 7 with stride 1 that can map to 57, 55, 53 possible informative patches for the input sentence with the normalized spatial length of 59. The three localization branches are named as $B^{3}$, $B^{5}$ and $B^{7}$, respectively.
	
	Before classification for each branch in AnchorNet-I, we utilize a linear $1\times 1$ convolution to encode the representations into a 1000-channel feature map and combine it with the spatial attention map by broadcast element-wise multiplication to a class-specific activation map $\mathbf{F}^{j}\in \mathbb{R}^{H^{j}\times W^{j}\times 1000}$, where $j\in\{63,95,111\}$ denotes the given branch, $H$ and $W$ denote the spatial height and width, respectively. Each channel of $\mathbf{F}^{j}$ denotes an activation map of the corresponding class generated by branch $B^{j}$. And then we apply a global average pooling for $\mathbf{F}^{j}$ and a softmax layer to obtain the class probability distribution. The outputs of all branches after softmax layer are supervised by the cross-entropy loss with image-level labels and without any bounding box annotations. The process of AnchorNet-T is similar with AnchorNet-I except that the spatial size of feature is 1D instead of 2D, as illustrated in Figure \ref{AnchorNet-T}.
	
	Each branch is attached with an attention branch, which assists spatially feature localization. An additional supervision is introduced for attention features, which can be much easier to learn discriminative features and facilitate the attention localization. The connection to GAP and FC is only used for auxiliary training, and thus having no effect at inference. The overall architecture of attention branch is shown in supplementary material section B.

	\begin{figure}[tbp]  
		\centering  
		\includegraphics[width=1.0\linewidth]{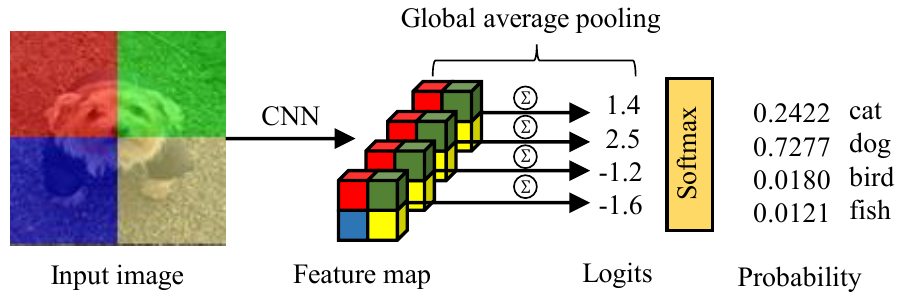}
		\caption{The diagram of linear aggregation for patch feature. For better understanding, we separate the input image into a grid of non-overlapping patches with $112\times 112$ size.}  
		\label{lienar}
	\end{figure}

	\subsubsection{Why linear aggregation preserves interpretability} Compared with the setting of popular fully-connected (FC) layer, it is noteworthy that we just perform a linear average aggregation along the spatial dimension for a 3D feature map $\mathbf{F}^{j}$ and then attach softmax function. After end-to-end training, each channel of $\mathbf{F}^{j}$ emphasizes class-specific informative locations, and activation value can evaluate the importance of location for classification result. As a result, if we would like to pinpoint exactly how various patches contribute the final prediction, the critical problem can be equal to how each patch exactly maps the corresponding high-level location. One simple yet effective method is generating a 3D class-specific feature map until GAP layer by CNN extractor and implement the linear aggregation for classification, as shown in Figure \ref{lienar}. As we all know, convolutional filter models pixel relationships in a local neighborhood, so each location of the final 3D feature map extracted by accumulated convolutions integrates a fixed part of input image, and we can exactly map each location to the corresponding patch via the calculation of accumulated RF, as mentioned in section \ref{Review}. And we further discard FC layer because  it can facilitate the interaction between patch-wise evidences, though improving the performance of classification, it may destroy the interpretability of exact mapping.

	\subsection{Localizing Multi-scale Informative Patches}
	Given an input image $x$ to AnchorNet-I, localization branch $B^{j}$ can predict the class probability distribution $[B^{j}_{1}(x),B^{j}_{2}(x),...,B^{j}_{1000}(x)]$, where $B^{j}_{y}(x)$ denotes the probability of the class $y$, $y\in \{1,2,...,1000\}$, $j\in \{63,95,111\}$. Then we make a simple classification decision for individual branches as following:
	\begin{equation}
	\label{decision}
	Y^{j}=\arg \max_{y}B^{j}_{y}(x),\ P^{j}=\max_{y}B^{j}_{y}(x)
	\end{equation}
	Where $Y^{j}$ and $P^{j}$ denote the predicted class and its probability by branch $B^{j}$, respectively. Then we can implement the systematic decisions of final class $\gamma$ and which branch (denoted as $B^{\theta}$) is used for patch localization according to (\ref{Y}) and (\ref{B}) as following, where equation (\ref{I}) is used for judging whether it is equal between two variables.
	\begin{equation}
	\label{I}
	I(a,b)=\left\{\begin{matrix}
	1, if \ a=b
	\\ 0, otherwise
	\end{matrix}\right.
	\end{equation}
	
	\begin{equation}
	\label{Y}
	\gamma=\left\{\begin{matrix}
	y, if \ \exists y,\sum_{j}(I(y,Y^{j}))\geqslant 2
	\\ Y^{\arg \max_{j}P^{j}}, otherwise
	\end{matrix}\right.
	\end{equation}
	\begin{equation}
	\label{B}
	B^{\theta}=B^{\arg \max_{j}[(I(P^{j},B^{j}_{\gamma}(x)))\cdot P^{j}]}
	\end{equation}
	\begin{algorithm}[tb]
		\caption{Localizing Informative Patches (LIP)}
		\label{alg:algorithm}
		\textbf{Input}: input image $\mathbf{I}$, heatmap $\mathbf{M}^{\theta}\in \mathbb{R}^{H^{\theta}\times W^{\theta}}$\\
		\textbf{Parameter}: maximum number of selected patches $K^{\theta}$, $IoU$ threshold $T$, percentage of coverage $\mathcal P^{\theta}$\\
		\textbf{Output}: collection of localized image patches $\mathcal S^{\theta}$
		
		\begin{algorithmic}[1] 
			\STATE $\mathcal{D}^{\theta}$=Reverse\_Sort(Flatten($\mathbf {M}^{\theta}$))
			\STATE 	$\mathcal S$=\{$\mathbf{p}$|$\mathcal{D}^{\theta}[1]\rightarrow\mathbf{p},\mathbf{p}\in \mathbf{I}$\}  \# Mapping the first index coordinate in $\mathcal{D}^{\theta}$ to the patch $\mathbf{p}$ in input image $\mathbf I$
			\FOR{$i=2:H\times W\times \mathcal P^{\theta}$}
			\STATE 	$\mathbf{p}_{candidate}=\mathbf{p}$|$\mathcal{D}^{\theta}[i]\rightarrow\mathbf{p},\mathbf{p}\in \mathbf{I}$
			\IF{$\forall s\in \mathcal S^{\theta}$,$IoU$($\mathbf{p}_{candidate}$, $s$)$<$$T$}
			\STATE $\mathcal S^{\theta}$ = $\mathcal S^{\theta} \cup $\{$\mathbf{p}_{candidate}$\}
			\ENDIF
			\IF{len($\mathcal S$) $==K^{\theta}$}
			\STATE \textbf{return} $\mathcal S^{\theta}$
			\ENDIF
			\ENDFOR
			
			\STATE \textbf{return} $\mathcal S^{\theta}$
		\end{algorithmic}
	\end{algorithm}

	Given the certain branch $B^{\theta},\theta\in\{63,95,111\}$, each channel of logits tensor $\mathbf{F}^{\theta}$ corresponds the  class-specific activation map, i.e. for the predicted class label $\gamma$, the heatmap $\mathbf{M}^{\theta}\in \mathbb{R}^{H^{\theta}\times W^{\theta}}$ can be obtained that is equal to $\mathbf{F}^{\theta}_{:,:,\gamma}$, which represents the interpretable contribution of each mapped patch for predicted class $\gamma$. 
	
	Instead of simply selecting top $K$ patches with maximum activations, we perform LIP in Algorithm \ref{alg:algorithm} to ensure the localized patches that are not only informative but also partly separated to cover more information. First, we flatten the $\mathbf{M}^{\theta}$ to a candidate index coordinates set $\mathcal{D}^{\theta}=\{(h,w)|h\in\{1,2,...,H^{\theta}\},w\in\{1,2,...,W^{\theta}\}\}$, and then sort them from maximum to minimum according to their corresponding activation values. Initially, we straightforward map the first coordinate point which has the maximum activation to the corresponding patch, mapping rule is as mentioned in section \ref{Review}, and put it in the collection $\mathcal S^{\theta}$. Next, we visit each index coordinate sequentially from front to back, the mapped patch with the size of $\theta \times \theta$ of which can be put in the $\mathcal S^{\theta}$ only if it can meet the following conditions: the $IoU$ of this patch between any patches in $\mathcal S^{\theta}$ is less than the threshold $T$. Where $IoU$ is a quite practical indicator to quantify the intersection between two patches A and B:
	\begin{equation}
	\label{iou}
	IoU=\left | A\cap B \right | / \left | A\cup B \right |
	\end{equation}
	Where $\left | \cdot \right |$ calculate the pixel number of the region. That means that localized patches can be controlled to be separated and informative concurrently by introducing the $IoU$ mechanism. When the number of patches in $\mathcal S^{\theta}$ achieves the upper limitation $K^{\theta}$, the final collection of patches can be obtained.
	
	Instead of performing LIP algorithm for text localization, we straightforward select the top-1 patch derived from the class-specific activation maps among $\mathbf{F}^{3}\in \mathbb{R}^{57\times 2}$, $\mathbf{F}^{5}\in \mathbb{R}^{55\times 2}$ and 
	$\mathbf{F}^{7}\in \mathbb{R}^{53\times 2}$, all of them are tagged in Figure \ref{AnchorNet-T}.
	
	\section{Experiments}
	\begin{figure*}[tbp]  
		\centering  
		\includegraphics[width=0.95\linewidth]{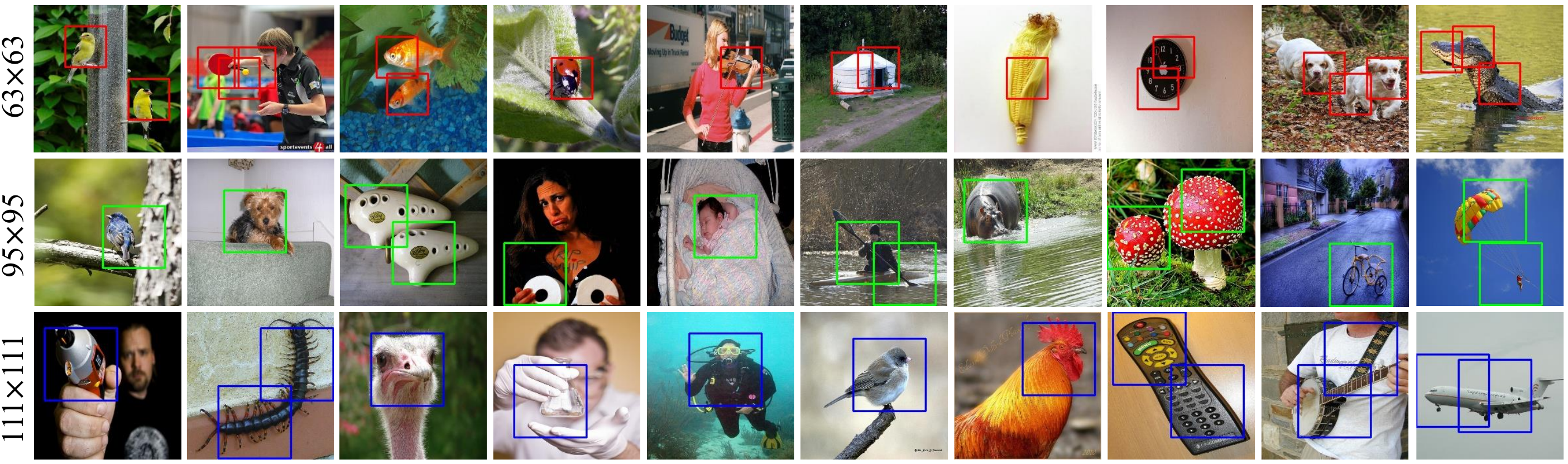}
		\caption{Examples of multi-scale informative image patches localized by AnchorNet-I. The first, second and third row denote the results localized by $B^{63},B^{95},B^{111}$, respectively. Note that each input image is adaptively assigned to one of three branches for localization according to its object property. More localized patches are provided in supplementary material section D.}  
		\label{patches}
	\end{figure*}
	\subsection{Dataset and Settings}
	We experiment AnchorNet on ImageNet (ILSVRC 2012) \cite{deng2009imagenet} and MR \cite{DBLP:conf/acl/PangL04} datasets  to validate the effectiveness of localizing multi-scale informative image and text patches. ImageNet is a large-scale and diverse dataset for image recognition, which contains 1.2 million training images and 50k validation images with 1000 classes, and includes both coarse- and fine-grained class distinction, e.g. over 100 fine-grained classes of dogs. MR contains 10662 movie reviews with positive/negative labels for sentiment recognition, where per review is one sentence. Training and hyperparameters settings are discussed in supplementary material section C.
	\begin{figure}[htbp]  
		\centering  
		\includegraphics[width=1.0\linewidth]{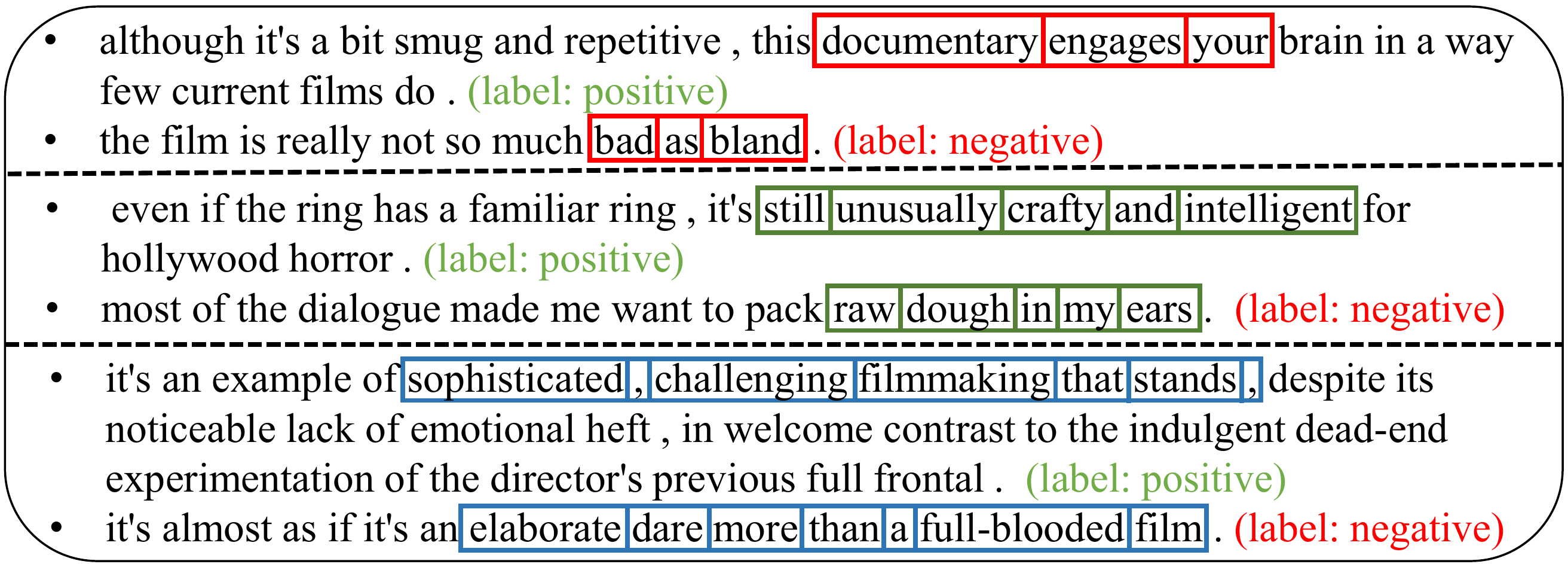}
		\caption{Examples of multi-scale informative text patches localized by AnchorNet-T. The first, second and third blocks denote the results localized by $B^{3},B^{5},B^{7}$, respectively.}  
		\label{text}
	\end{figure}
	
	\begin{figure}[htbp]  
		\centering  
		\includegraphics[width=1.0\linewidth]{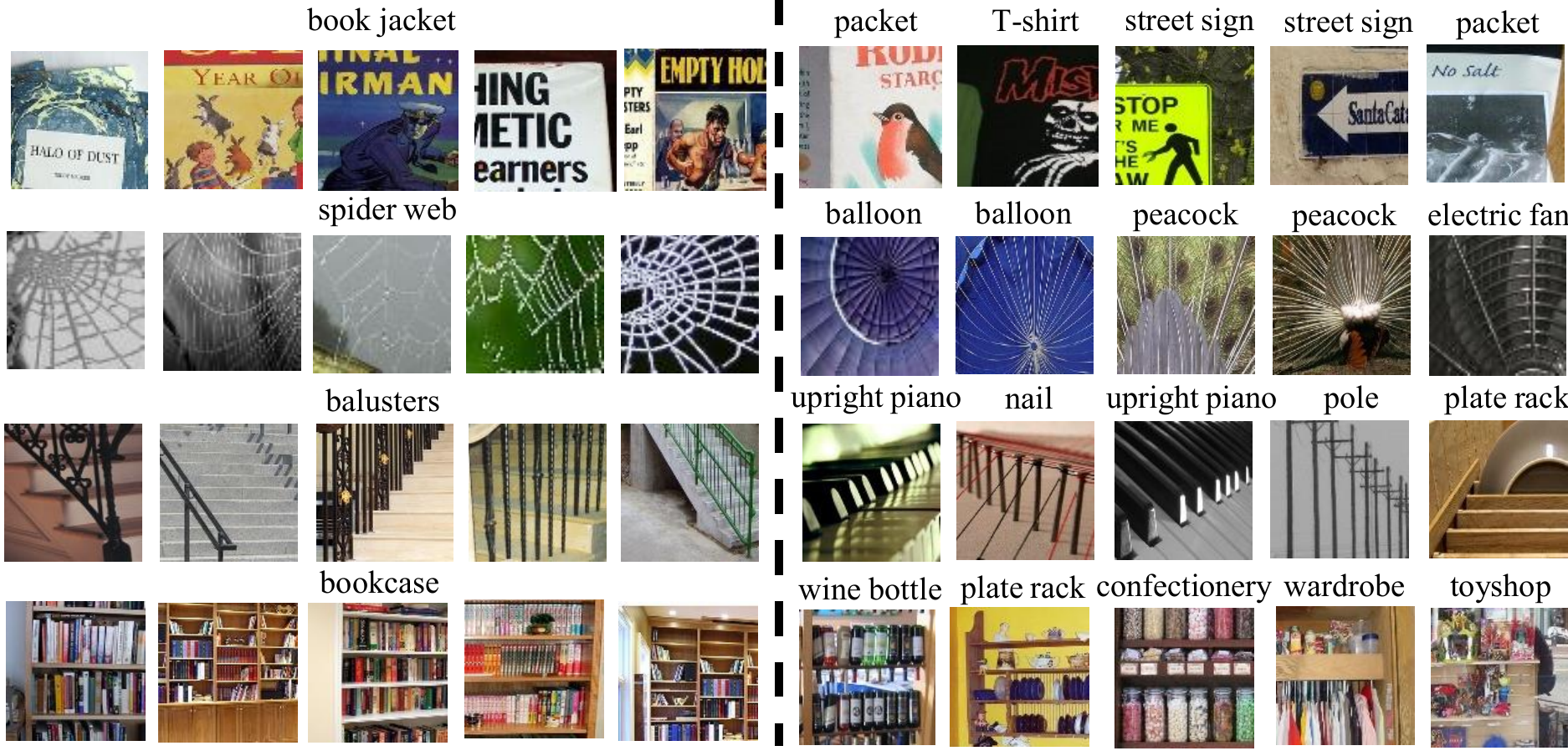}
		\caption{Informative patches for the given class from images with the same class (\emph{left}) and different classes (\emph{right}). Each row denotes one class corresponding its representative patches (\emph{left}) and confusion patches (\emph{right}).}  
		\label{confusion_patch}
	\end{figure}

	\begin{figure}[htbp]  
		\centering  
		\includegraphics[width=1.0\linewidth]{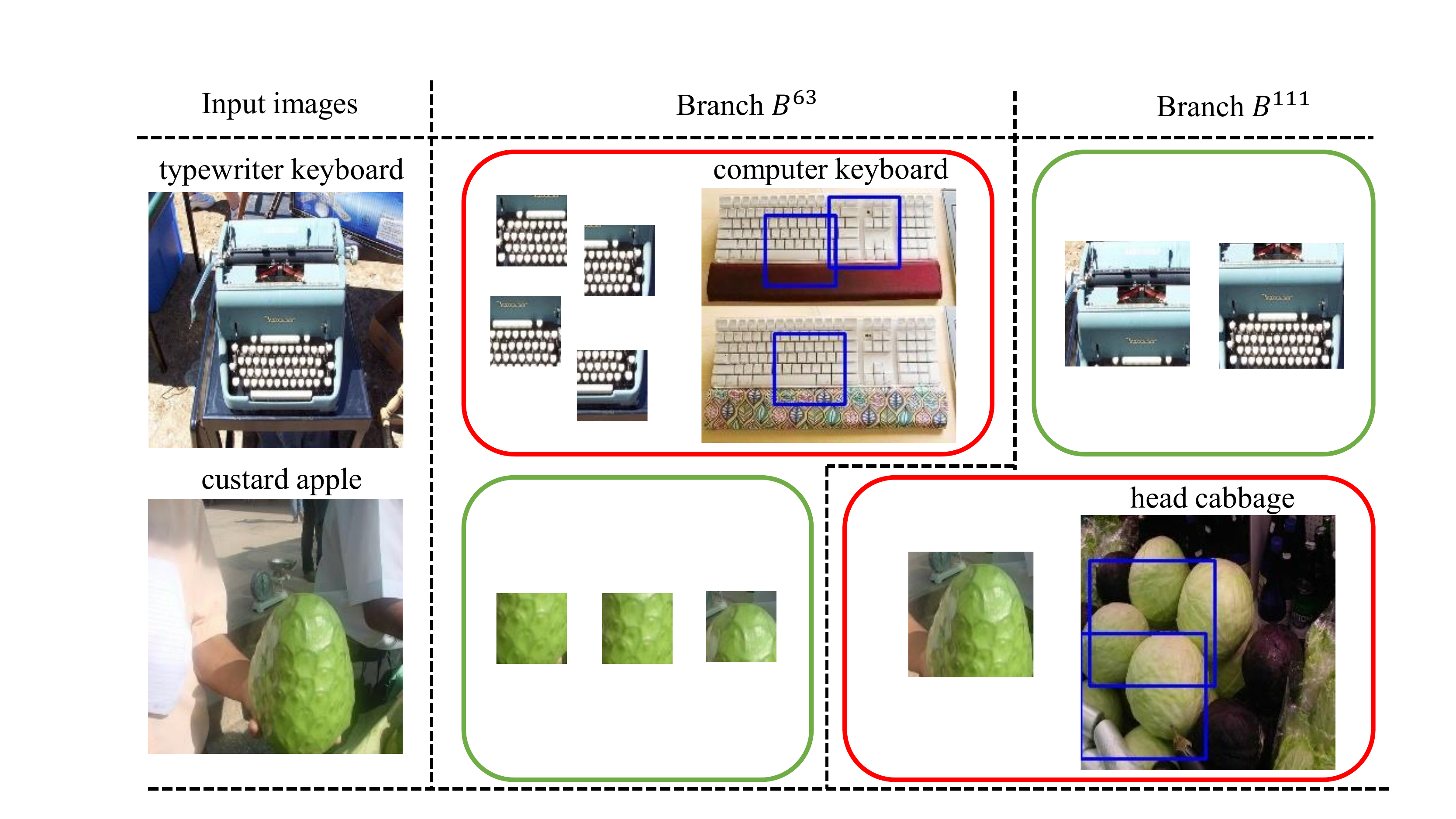}
		\caption{Misclassified cases of branch $B^{63}$ and $B^{111}$. For each branch, we show the localized patches with the most class evidences in a box, where the green and red margins indicate correct and incorrect predictions, respectively. In the cases of red box, we further show the representative image of misclassified category and localize the most relevant patches for classification by the corresponding branch.}  
		\label{confusion}
	\end{figure}

	\subsection{Localized Multi-scale Informative Patches}
	Visualization of localized multi-scale informative image patches is shown in Figure \ref{patches}. Somewhat surprisingly, AnchorNet-I only derived from classification task can be on a par with the method of object detection in localization performance.  Benefit from various RFs, AnchorNet-I is equipped with the capability of  multi-scale localization, where the branch with wider RF size is more prone to localize larger object and coarse-grained global features, such as scuba diver, remote control and airliner, which occupy the most part in images, are captured by $B^{111}$. While the branch with relatively narrower RF always localizes smaller object and fine-grained local features, e.g., $B^{63}$  not only captures the miniature object such as ladybug, fish and violin, but also identifies the local texture features of large objects, such as corn and crocodile. 
	
	Visualization of localized multi-scale informative text patches is shown in Figure \ref{text}. It can be observed that AnchorNet-T can adaptively capture the crucial evidences for sentiment recognition with various RFs (filter windows). Moreover, we argue that wider RF can generally capture longer range dependence of the contextual words, while narrower RF is also preferable when the decisive dependence is short-term.
	
	We would like to interpret why the misclassification happens in terms of the informative patches. We select a given class and perform AnchorNet-I across all validation images to find the representative patches from that class, and some confusion patches from other classes that provide informative evidences for the given class, as shown in Figure \ref{confusion_patch}. Visualization of these patches provide some insights on misclassification: e.g., book jacket features the design and text as evidences, which are also appeared on packet, t-shirt and street sign. The texture of spider web is quite similar with that of balloon, peacock's tail and electric fan. Balusters is always together with stairs, their shapes make them confuse with upright piano, nail, pole and plate rack. The layer-by-layer way of densely arranged books sometimes looks like as the similar as the arranged scenes of wine bottle, confectionery and wardrobe.

	Another intriguing case is the misclassification that may take place in both $B^{63}$ and $B^{111}$ due to their characteristics of feature extraction, as illustrated in Figure \ref{confusion}. Combined with the above discussion, we further consider that although narrow RF can capture local features, it may ignore the more informative global features, e.g., $B^{63}$ concentrates on local keyboard yet omits the global typewriter, leading to confusion with computer keyboard. In contrast, wide RF prefers localizing coarse-grained features but ignore local fine-grained features, e.g., the outline and color of custard apple are interpreted as evidences for head cabbage by $B^{111}$, which omit the different texture information between them, these decisions looks comprehensible.
	
	\begin{figure}[tbp]
		\centering  
		\includegraphics[width=1.0\linewidth]{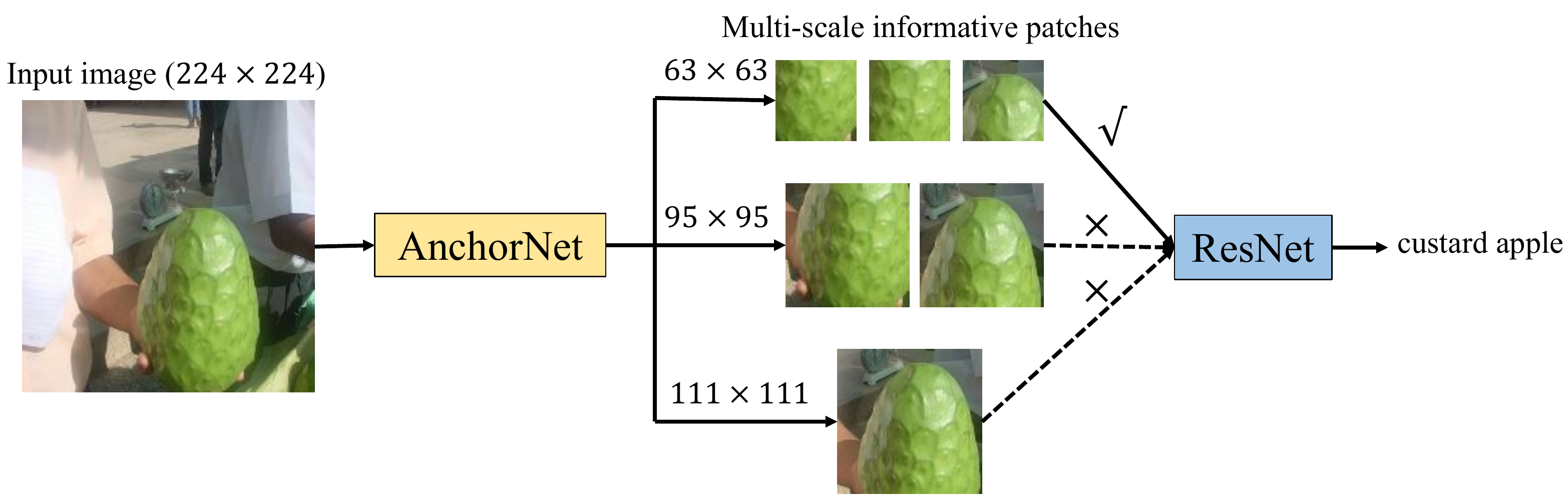}
		\caption{Pipeline of using informative patches localized by upstream AnchorNet for downstream classification model. }  
		\label{cls_framework}
	\end{figure}
	
	\begin{table}[t]
		\centering
		\caption{Comprehensive performance of multi-scale informative image patches for classification on ImageNet. Each bold entry denotes the overall result by weighted average of three branches, where each weight corresponds the proportion of the number of localized images.}
		\begin{tabular}{c|c|c|c|c|c|c|c}
			\hline
			& & \multicolumn{2}{c|}{}&\multicolumn{2}{c|}{}&\multicolumn{2}{c}{} \\[-9pt]
			
			Model &Scale&\multicolumn{2}{c|}{FLOPs} &\multicolumn{2}{c|}{Top-1 (\%)}&\multicolumn{2}{c}{Top-5 (\%)} \\\hline
			
			& &\multicolumn{2}{c|}{} &\multicolumn{2}{c|}{}&\multicolumn{2}{c}{} \\[-9pt]
			
			\multirow{4}{*}{ResNet} &$224^{2}$&  \multicolumn{2}{c|}{1x}&  \multicolumn{2}{c|}{72.6} & \multicolumn{2}{c}{91.0}\\  \cline{2-8}
			& & && &&&\\[-9pt]
			
			&$63^{2}$&0.46x&\multirow{3}{*}{\textbf{0.51x}}& 68.6  &\multirow{3}{*}{\textbf{71.1}}  & 87.2   &\multirow{3}{*}{\textbf{89.1}} \\   
			&$95^{2}$&  0.54x&& 72.1    & & 89.9    & \\  
			&$111^{2}$& 0.54x&&72.2 &&89.9    &   \\ \hline

			& &\multicolumn{2}{c|}{} &\multicolumn{2}{c|}{}&\multicolumn{2}{c}{} \\[-9pt]
			\multirow{4}{*}{ResNeXt} &$224^{2}$& \multicolumn{2}{c|}{1x}&  \multicolumn{2}{c|}{75.7} & \multicolumn{2}{c}{92.8}\\  \cline{2-8}
			& & && &&&\\[-9pt]
			&$63^{2}$& 0.47x&\multirow{3}{*}{\textbf{0.51x}}& 72.8    &\multirow{3}{*}{\textbf{74.1}}  & 89.8 &\multirow{3}{*}{\textbf{90.7}} \\   
			&$95^{2}$& 0.53x& & 74.3    & &91.0    & \\  
			&$111^{2}$& 0.56x& & 75.0    && 91.3   &   \\ \hline 
			& & \multicolumn{2}{c|}{}&\multicolumn{2}{c|}{}&\multicolumn{2}{c}{} \\[-9pt]
			
			\multirow{4}{*}{DenseNet} &$224^{2}$& \multicolumn{2}{c|}{3.4}& \multicolumn{2}{c|}{74.8} & \multicolumn{2}{c}{92.5}\\  \cline{2-8}
			
			& & && &&&\\[-9pt]
			&$63^{2}$& 0.47x&\multirow{3}{*}{\textbf{0.50x}}&68.6 &\multirow{3}{*}{\textbf{71.5}}  &88.1 &\multirow{3}{*}{\textbf{89.8}} \\   
			&$95^{2}$& 0.53x& &72.3&   &90.2   & \\  
			&$111^{2}$& 0.53x& &73.3  && 90.8 &   \\ \hline

		\end{tabular}
		
		\label{cls}
	\end{table}

	\begin{table}[t]
		\centering
		\caption{Localized multi-scale informative text patches for classification on MR.}
		\begin{tabular}{ccc}
			\toprule
			Model & FLOPs & Accuracy (\%)
			\\\midrule
			CNN &1x& 83.4\\ 
			AnchorNet-T+CNN &0.07x& 81.7\\
			\bottomrule 
		\end{tabular}
		
		\label{mr_cls}
	\end{table}
	
	%
	
	\subsection{Using Semantic Patches for Classification}

	We further conduct downstream classification according to localized patches so as to verify their representations for semantics of the original images, the pipeline of which is shown in Figure \ref{cls_framework}. As shown in Table \ref{cls}, we utilize ResNet-50 \cite{he2016deep}, ResNeXt-50 \cite{xie2017aggregated}, and DenseNet-169 \cite{huang2017densely} fine-tuned on training patches to implement classification tasks. FLOPs denotes the average number of floating point operations for processing one validation image, which refers to the initial image if $224\times 224$ scale, otherwise the all corresponding localized patches. We report top-1 and top-5 accuracy to measure the performance of classification. The results are evaluated on ImageNet validation set, where each image is localized by one of three branches using LIP algorithm. Across all 50K validation images, from where 15050, 18047, 16903 images are localized by $B^{63}$, $B^{95}$ and $B^{111}$ corresponding with 5.6, 2.9 and 2.1 patches for an image on average, respectively. Since one image may generate multiple relevant patches, we implement the final decision by adding the softmax distributions of them, and determine the class with maximum probability. Table \ref{cls} shows that without any changes of models, using multiple semantic patches instead of the original images can achieve about $2\times $ acceleration with tiny drop of top-1 accuracy, varying from $1.5$\% on ResNet-50 as minimum to $3.3$\% on DenseNet-169 as maximum, which demonstrates the multi-scale patches localized by AnchorNet-I can retain the most semantics while significantly speeding up the inference. We further demonstrate the performance of remarkable acceleration and good accuracy is attributed to localizing but unable to be obtained by vanilla scale reduction from the original images in supplementary material section E.
	
	Similar experiment is also conducted on text localization to evaluate the performance of AnchorNet-T.  After running 1068 sentences in test set, 454, 547 and 67 sentences are localized by $B^{3}$, $B^{5}$ and $B^{7}$, respectively. Per sentence corresponds one patch. We utilize CNN \cite{DBLP:conf/emnlp/Kim14} as the downstream model for text patches classification behind AnchorNet-T, as shown in Table \ref{mr_cls}. It can be observed that localized informative patches can retain the most evidences for sentiment recognition with only few words, thus resulting in the significantly reduction of FLOPs. Moreover, we argue that although $B^{7}$ could capture longer range dependence, AnchorNet-T tends to use $B^{3}$ and $B^{5}$ to localize patches, which indicates that continuous 3 or 5 words are sufficient for sentiment recognition on MR.
	\subsection{What have attention branches learned?}
	\begin{figure}[tbp]
		\centering  
		\includegraphics[width=1.0\linewidth]{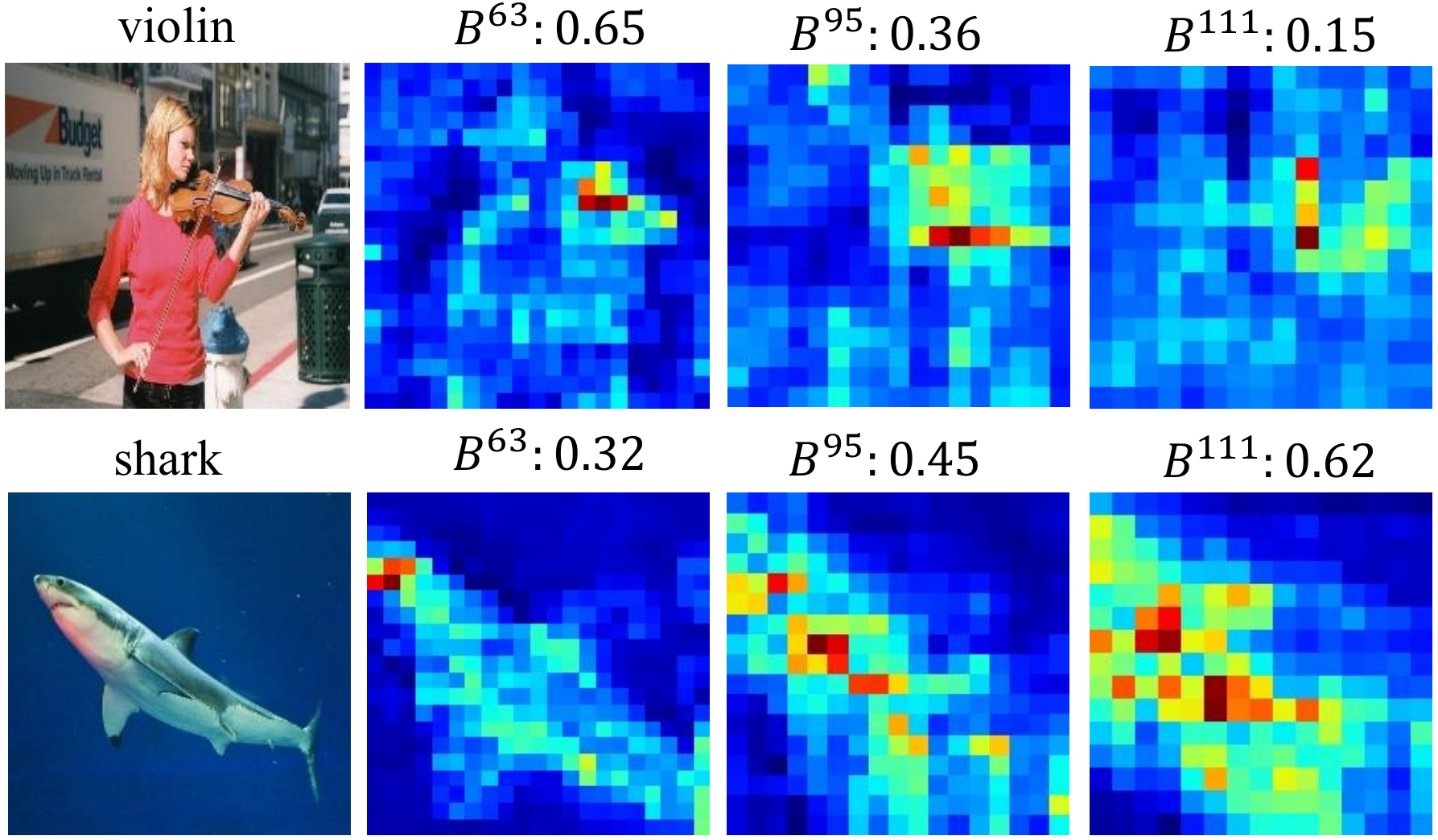}
		\caption{Visualization of spatial attention maps, which are generated by attention branches of $B^{63},B^{95},B^{111}$ from left to right conditioned on the corresponding input images.}  
		\label{attentions}
	\end{figure}
	
	It is universally known that spatial attention map always emphasizes the most salient regions and could provide semantic information. In AnchorNet, attention branches are introduced to assist semantic feature localization, we further visualize the generated spatial attention maps from various branches along with the predicted probability, which are depicted in Figure \ref{attentions}. It can be observed that all attention branches can roughly attend to the informative locations, while ignoring the backgrounds. Additionally, it can be seen clearly that narrow RF mainly concentrates on the details and high frequency information, while wide RF focuses more on holistic property and low frequency information, due to the various grains on feature scanning. In terms of decision-making, we argue that the localization branch would generally infer a higher confidence if the scale of input object is most suitable for its RF compared with other branches. Specifically, the RF of $63\times 63$ exactly pinpoints the small violin, whereas other RFs aggregate the information of both object and noise background, which may affect the effective classification. On the other hand, wider RF captures the overall shape of shark, thus providing a solid class evidence, while narrow RF is weak in building wide context dependencies, thus leading to a more uncertain in decision.
	%
	
	%
	%
	%
	%
	\begin{figure}[tbp]  
		\centering  
		\includegraphics[width=1.0\linewidth]{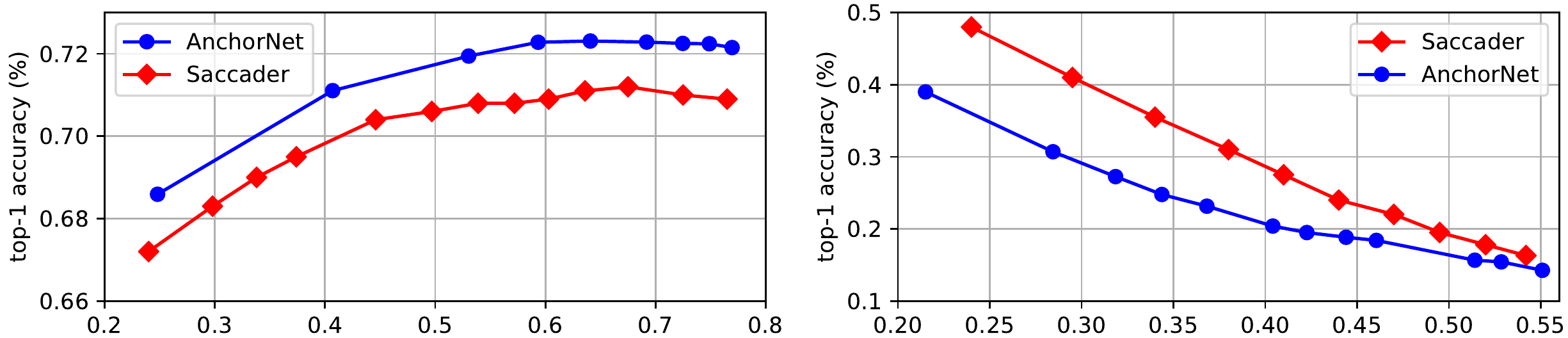}
		\caption{Relationships between accuracy (evaluated on downstream ResNet-50) and covered area (\emph{left}), or masked area (\emph{right}) of patches localized by AnchorNet-I and Saccader.}  
		\label{covered}
	\end{figure}
	
	\subsection{Comparison with State-of-the-art}
	
	\begin{figure}[tbp]  
		\centering  
		\includegraphics[width=1.0\linewidth]{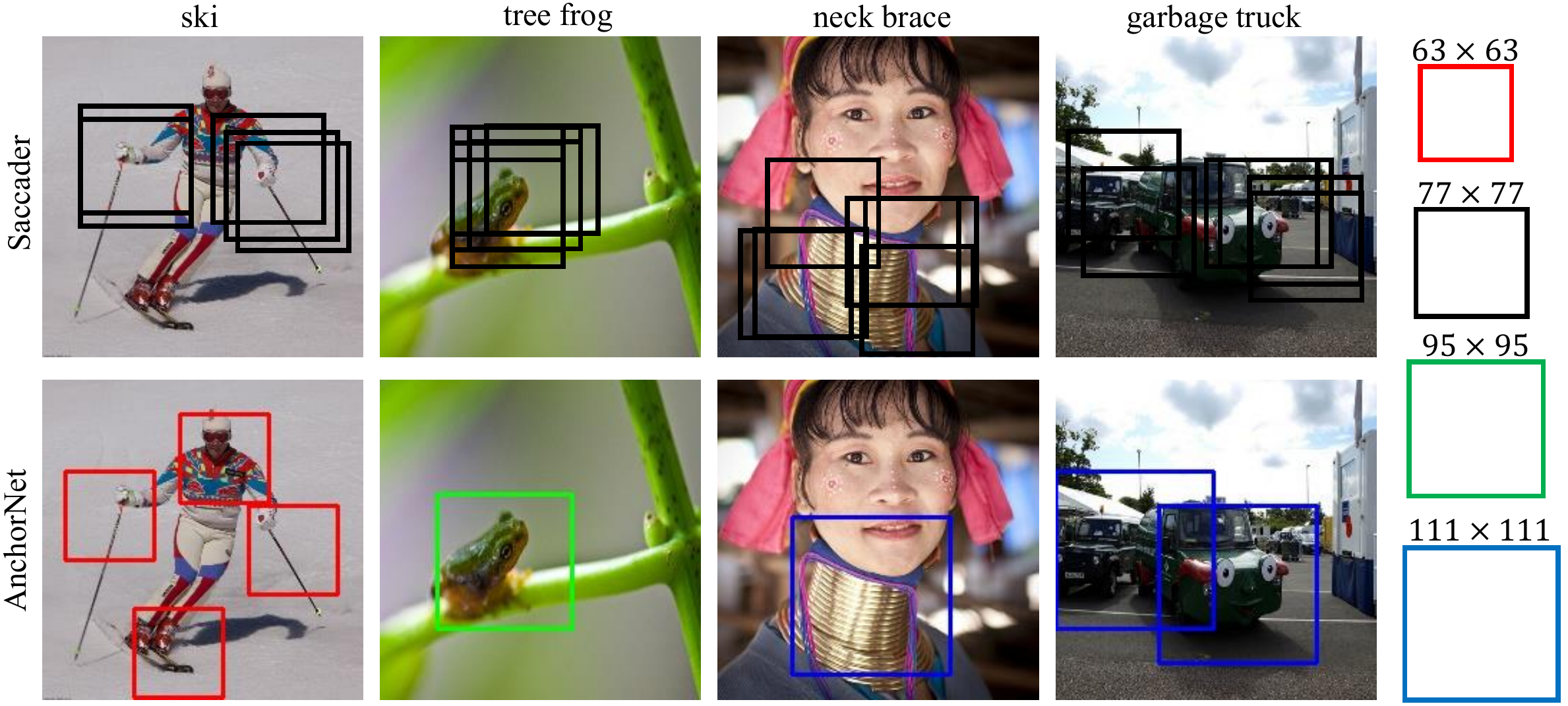}
		\caption{Comparison of AnchorNet-I against Saccader about localized informative image patches.}  
		\label{visual_saccader}
	\end{figure}
	
	Figure \ref{covered} quantitatively compares our AnchorNet-I with  state-of-the-art Saccader \cite{elsayed2019saccader} on ImageNet according to the performance of downstream classification by their localized patches, it can be observed that both of them generally lead to better accuracy as the covered area increases. Moreover, using relevant patches localized by AnchorNet-I can achieve the better performance than Saccader under the same coverage, which proves the superiority of AnchorNet-I for downstream classification task is not simply attributed to wider image coverage. We further investigate the importance of localized patches by using them to mask the original images (i.e., set the pixels to 0) and then perform a classification on resulting images. Figure \ref{covered} shows that masking by AnchorNet-I leads to more significant drop in performance than Saccader, especially when the masked coverage is small, which demonstrates that patches localized by AnchorNet-I have more valuable information for image recognition. Moreover, AnchorNet-I (parameters: 1.6M, FLOPS: 0.5G) only uses fewer complexity than Saccader (parameters: 33.6M, FLOPS: 21.6G) by an order of magnitude.
	
	We further explain why AnchorNet-I outperforms Saccader by visualizing their localized informative patches in Figure \ref{visual_saccader}. We can observe that one of prominent superiorities of AnchorNet-I is attributed to the capability of multi-scale localization, which can adaptively capture the object according to its scale using various RFs, while Saccader only has a single RF thus inducing a ineffective modeling on multi-scale. Moreover, AnchorNet-I uses soft attention mechanism and  thus suffers easier pixel-wise optimization than Saccader, which leads to better performance in localizing local informative features, e.g. AnchorNet-I captures the object of ski and its associated poles, glasses and hat, while Saccader only captures secondary poles. Benefit from LIP algorithm, patches localized by AnchorNet-I are always partly separated so that they can efficiently cover more semantic information than that of Saccader. 
	
	\subsection{Robustness to Noise Image}
	\begin{figure}[tbp]
		\centering  
		\includegraphics[width=0.95\linewidth]{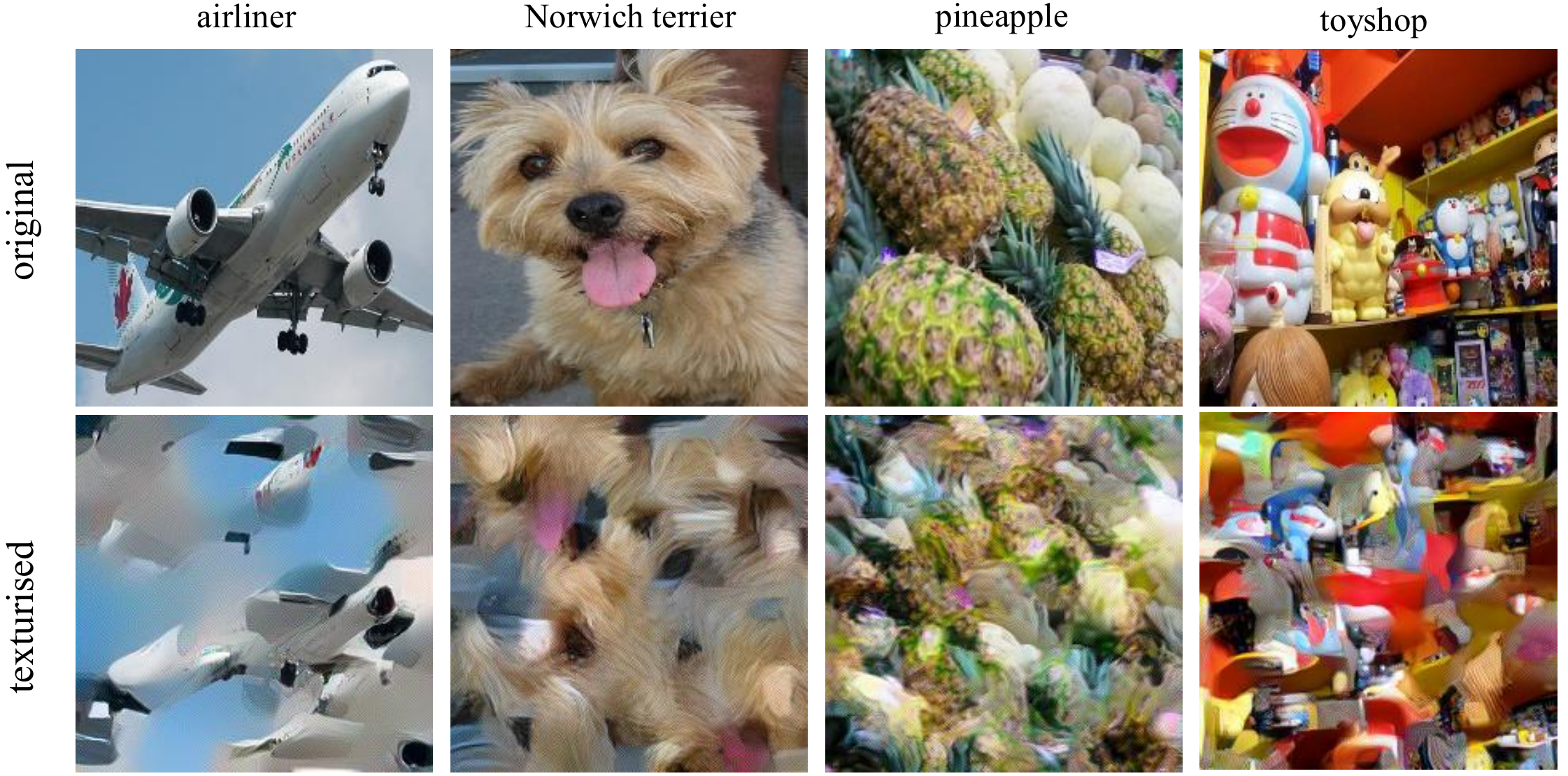}
		\caption{Examples of texturised images. Deep CNN classifier can still reach a good performance on these texturised images, while humans greatly lose the recognition ability. }  
		\label{texturised}
	\end{figure}
	Texture synthesis \cite{DBLP:journals/corr/GatysEB15a} of the image preserves the local spatial features while scrambling the global spatial arrangement, as shown in Figure \ref{texturised}. The texturised image is generated by style transfer to a white noise image using VGG-19 \cite{DBLP:journals/corr/SimonyanZ14a}. We can observe that these texturised images significantly increase the difficulty of the object recognition for humans. Theoretically, texturised images would not affect the feature localization and classification decision by AnchorNet-I, because each patch is independent with other patches for the final decision, which is unrelated to global patch arrangement. Experiment across the texturised ImageNet validation set leads to little impact on downstream classification accuracy (from 71.1 to 62.5) compared with the severely degradation of humans, which demonstrates that both feature localization and downstream decision are robust to texture synthesis. We also show that AnchorNet-I is robust to adversarial attack in supplementary material section F.

	\section{Conclusion}
	We construct an interpretable CNN framework named AnchorNet to provide patch-wise evidences derived from  classification task, meanwhile localizing multi-scale informative patches. Experiments on image and text localization show that multi-scale informative patches  retain the most semantics and evidences of the original media, while accelerating the inference process for downstream tasks. Theoretically, the framework of AnchorNet can be extended any media with CNN-based models, e.g. audio and speech recognition \cite{DBLP:conf/icassp/HersheyCEGJMPPS17,DBLP:conf/icassp/ZhangCJ17}, which will become our future works. We hope our AnchorNet may inspire the future study of interpretable informative feature localization and application.   
	
	\bibliographystyle{ACM-Reference-Format}
	\bibliography{sample-base}
	
	\appendix
	\section{AnchorNet-T}
	The structural details of AnchorNet-T including head and localization branches are shown in Table \ref{head} and \ref{tails}, where IR denotes the spatial size of input, Operator denotes the type of convolution and its kernel size, In denotes the number of input channel, Out denotes the number of output channels, $s$ denotes the stride of convolution, RF denotes the accumulated RF size until the current layer. In Table \ref{tails}, the final row of each block denotes the convolution in attention branch.
	\begin{table}[h]
		\centering
		\caption{The structural settings of head.}
		\begin{tabular}{c|c|c|c|c|c}  
			\toprule
			IR & Operator & In &Out&$s$&RF \\
			\midrule
			59  & conv1d,1   &300  &32&0&1   \\
			59  & conv1d,1 & 32&64&0&1   \\
			\bottomrule
		\end{tabular}
		\label{head}
	\end{table}
	
	\begin{table}[ht]
		\centering
		\caption{The structural settings of localization branches.}
		\begin{tabular}{c|c|c|c|c|c}
			\toprule
			IR & Operator & In &Out&$s$&RF \\
			\midrule
			57& conv1d,3& 64&128&0&3   \\
			57& conv1d,1& 128&64&0&3   \\
			\midrule
			55& conv1d,5& 64&128&0&5   \\
			55& conv1d,1& 128&64&0&5   \\
			\midrule
			53& conv1d,7& 64&128&0&7   \\
			53& conv1d,1& 128&64&0&7   \\
			\bottomrule
		\end{tabular}
		\label{tails}
	\end{table}
	
	\section{Attention Branch}
	To assist feature learning for the localization branch in AnchorNet-I, we further construct an attention branch to emphasize informative locations by generating a spatial attention map. A bottleneck is applied to produce an additional feature map $\mathbf{X}\in \mathbb{R}^{H\times W\times C}$ from main branch for attention localization, where $H$ and $W$ denote the spatial height and width, $C$ denote the number of channels. Then a $1\times 1$ convolutional filter compacts $\mathbf{X}$ along the channel dimension to $\tilde{\mathbf{G}}\in \mathbb{R}^{H\times W\times 1}$, and followed by a softmax function to normalize spatial weights $\mathbf{G}\in \mathbb{R}^{H\times W\times 1}$:
	\begin{equation}
	\label{spatial}
	\mathbf{G}_{i,j,1}=\frac{e^{\tilde{\mathbf{G}}_{i,j,1}}}{\sum_{h=1}^{H}\sum_{w=1}^{W}e^{\tilde{\mathbf{G}}_{h,w,1}}}
	\end{equation}
	According to normalized spatial weights $\mathbf{G}$, we employ global weighted average pooling to $\mathbf{X}$ and produce a channel attention map $\tilde{\mathbf{C}}\in \mathbb{R}^{1\times 1\times C}$, the $c$-th channel of $\tilde{\mathbf{C}}$ is as (\ref{pool}), $*$ denotes the broadcast element-wise multiplication here. Channel attention map $\tilde{\mathbf{C}}$ can capture channel-wise dependencies, and can be considered as the importance of each channel.
	\begin{equation}
	\label{pool}
	\tilde{\mathbf{C}}_{c}=\sum_{h=1}^{H}\sum_{w=1}^{W} \mathbf{X}_{h,w,c}*\mathbf{G}_{h,w,1} 
	\end{equation}
	Then softmax function normalizes the $\tilde{\mathbf{C}}$ to generate the final \textbf{channel attention map} $\mathbf{C}\in \mathbb{R}^{1\times 1\times C}$:
	\begin{equation}
	\label{channel}
	\mathbf{C}_{1,1,c}=\frac{e^{\tilde{\mathbf{C}}_{1,1,c}}}{\sum_{i=1}^{C}e^{\tilde{\mathbf{C}}_{1,1,i}}}
	\end{equation}
	According to normalized channel weights $\mathbf{C}$, we employ weighted shrinking of $\mathbf{X}$ along the channel dimension to generate a spatial attention map $\tilde{\mathbf{S}}\in \mathbb{R}^{H\times W\times 1}$:
	\begin{equation}
	\label{un_spatial}
	\tilde{\mathbf{S}}_{i,j,1}=\sum_{c=1}^{C} \mathbf{X}_{i,j,c}*\mathbf{C}_{1,1,c} 
	\end{equation}
	After applying softmax function to $\tilde{\mathbf{S}}$, we output the final \textbf{spatial attention map} $\mathbf{S}\in \mathbb{R}^{H\times W\times 1}$ that we need:
	
	\begin{equation}
	\label{final_spatial}
	\mathbf{S}_{i,j,1}=\frac{e^{\tilde{\mathbf{s}}_{i,j,1}}}{\sum_{h=1}^{H}\sum_{w=1}^{W}e^{\tilde{\mathbf{s}}_{h,w,1}}}
	\end{equation}
	
	Spatial attention map $\mathbf{S}$ can capture the spatially pixel-wise importance, i.e. highlighting the informative region while suppressing uninformative region, thus providing a solid assistance for location localization. 
	
	The attention mechanism of AnchorNet-T is the same as that of AnchorNet-I with changing the spatial size from 2D to 1D. Moreover, we conduct ablation study on AnchorNet-T to validate the effectiveness of attention branch, the accuracy of three localization branches is reported in Table \ref{att_branch}. We can observe that attention branches consistently improve the performance across the three localization branches, and auxiliary training is a effective complementary method for optimization of the attention mechanism.
	\begin{table}[h]
		\caption{Ablation study of attention branch on AnchorNet-T.}
		\label{att_branch}
		\begin{tabular}{lccc}
			\toprule
			Method&$B^{3}$&$B^{5}$&$B^{7}$\\
			\midrule
			Baseline & 80.05& 79.78&79.87\\
			+attention &80.81& 80.99&80.90\\
			+attention+auxiliary training &\textbf{81.46}& \textbf{81.27}&\textbf{81.37}\\
			\bottomrule
		\end{tabular}
	\end{table}
	
	\begin{figure*}[htbp]
		\centering  
		\includegraphics[width=1.0\linewidth]{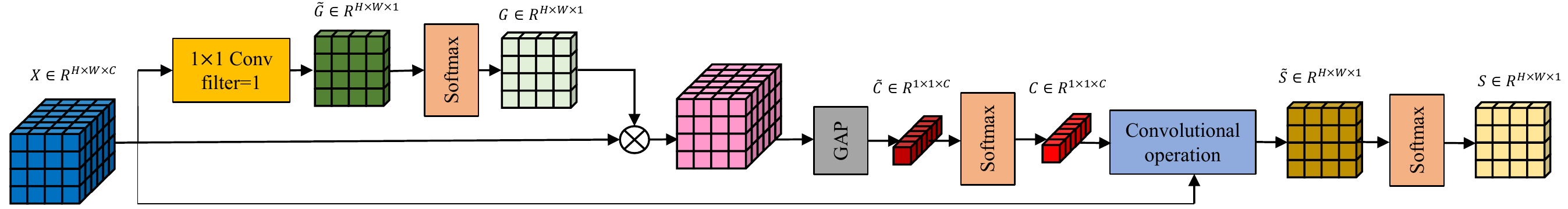}
		\caption{Illustration of the overall architecture of attention branch.}  
		\label{attbranch}
	\end{figure*}
	
	\begin{figure*}[htbp]
		\centering  
		\includegraphics[width=1.0\linewidth]{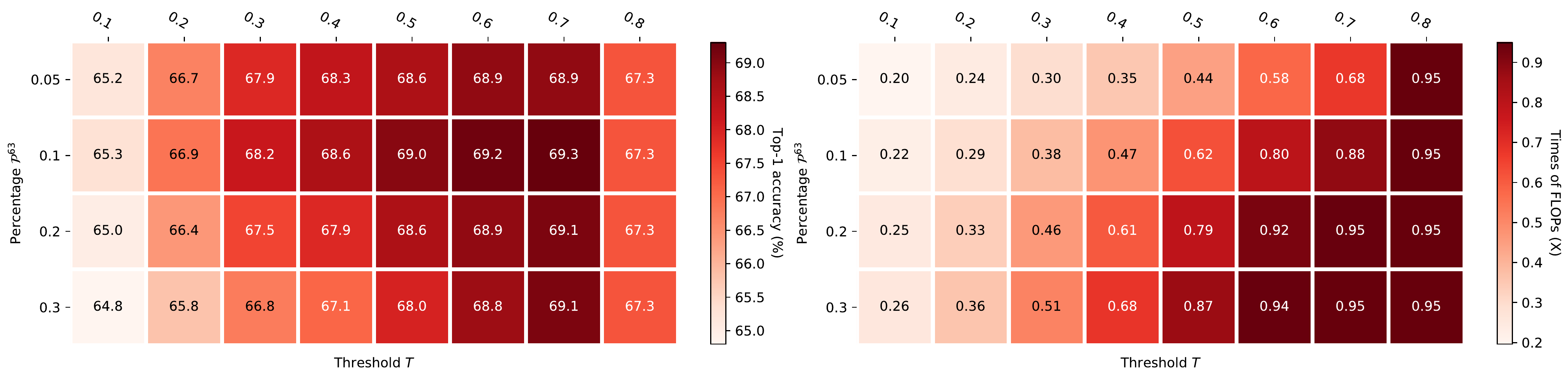}
		\caption{Top-1 accuracy (\emph{left}) and times of FLOPs (\emph{right}) compared with the original $224\times 224$ scale on ImageNet evaluated on downstream ResNet-50 across various combinations of percentage $\mathcal{P}^{63}$ and threshold $T$.}  
		\label{tune_63}
	\end{figure*}
	\begin{figure*}[htbp]
		\centering  
		\includegraphics[width=1.0\linewidth]{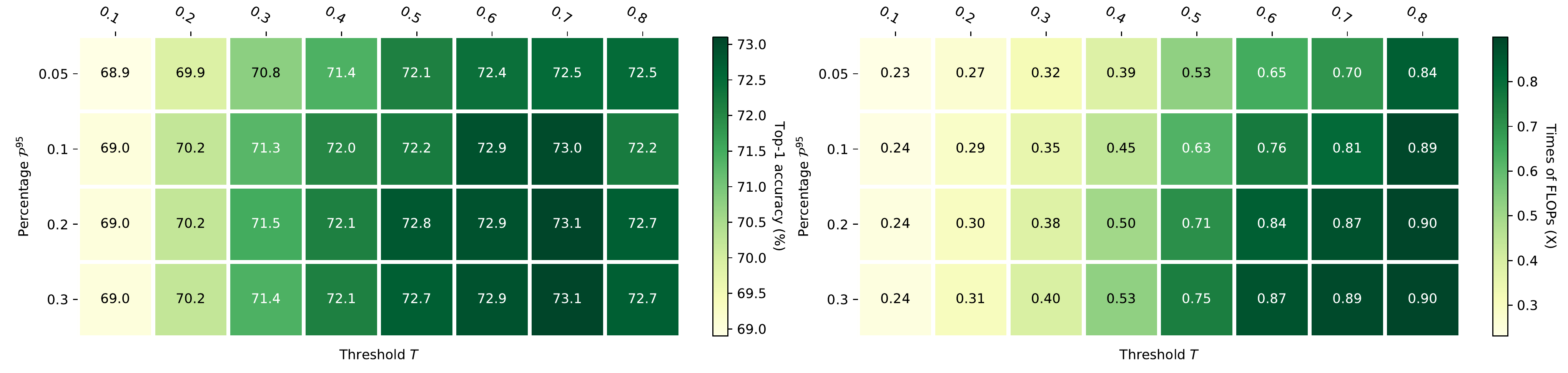}
		\caption{Top-1 accuracy (\emph{left}) and times of FLOPs (\emph{right}) compared with the original $224\times 224$ scale on ImageNet evaluated on downstream ResNet-50 across various combinations of percentage $\mathcal{P}^{95}$ and threshold $T$.}  
		\label{tune_95}
	\end{figure*}
	\begin{figure*}[htbp]
		\centering  
		\includegraphics[width=1.0\linewidth]{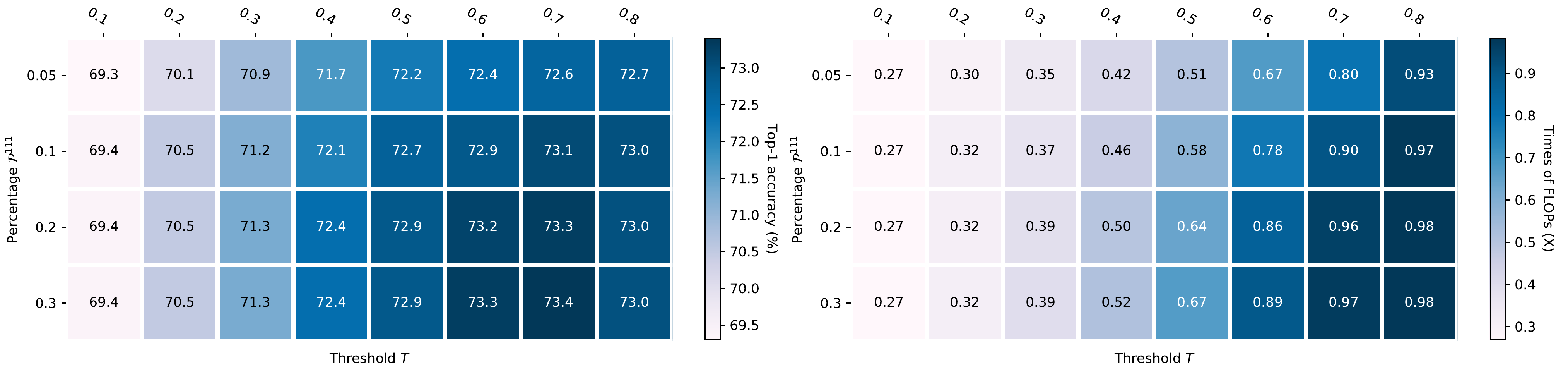}
		\caption{Top-1 accuracy (\emph{left}) and times of FLOPs (\emph{right}) compared with the original $224\times 224$ scale on ImageNet evaluated on downstream ResNet-50 across various combinations of percentage $\mathcal{P}^{111}$ and threshold $T$.}  
		\label{tune_111}
	\end{figure*}
	\section{Training and Hyperparameters settings}
	For Imagenet dataset, we conform the protocol in \cite{elsayed2019saccader} that using training set to train model and validation set to localize the informative patches.  At training time, standard data augmentation is employed following He \emph{et al.}~\shortcite{he2016deep}, and we use synchronous SGD with a momentum of 0.9 , a batch size of 256 and a weight decay of $10^{-4}$ for 100 epochs. The learning rate starts at 0.1 and decayed by a factor of 10 every 30 epochs. At test time, each image is resized to $224\times 224$ pixels on the validation set. 
	
	For MR dataset, we conform the protocol in \cite{DBLP:conf/emnlp/Kim14} that perform cross validation to divide the whole set into training set, validation set and test set with the proportions of 80\%, 10\% and 10\%, respectively. We use training set to train the model, validation set to adjust hyperparameters and test set to localize the informative patches. All words are initially embedded to pre-trained
	vectors from \emph{word2vec}\footnote{https://code.google.com/archive/p/word2vec/} with the dimensionality of 300. The embedded vectors are tunable across the course of training. We use Adadelta optimizer with an initial learning rate of 0.5 and a batch size of 50 for 50 epochs.

	About LIP algorithm, we set $K^{63}=24$, $K^{95}=10$, $K^{111}=8$, $T=0.8$, $\mathcal P^{63}=\mathcal P^{95}=\mathcal P^{111}=0.3$, for producing more training patches so as to fine-tune the downstream models and enhance its robustness for recognition tasks. For test stage, we adopt stricter settings of $K^{63}=12$, $K^{95}=5$, $K^{111}=4$, $T=0.5$, $\mathcal P^{63}=\mathcal P^{95}=\mathcal P^{111}=0.05$ for producing the most informative as well as a small number of patches to retain both performance and acceleration for downstream tasks.

	We further investigate the sensitivity of hyperparameters involved in LIP algorithm by evaluating the localized patches for downstream classifier ResNet-50. At inference stage, we would not like to the total area of localized patches exceeds the corresponding image, thus $K^{j}=\left \lfloor (224\times 224)/(j\times j) \right \rfloor, j\in\{63, 95, 111\}$. The results of top-1 accuracy and corresponding FLOPs across various combinations between percentage $\mathcal{P}^{j}$ and threshold $T$ are shown in Figure \ref{tune_63}, \ref{tune_95} and \ref{tune_111}. It can be observed that AnchorNet localizes more patches along with the increasing of $\mathcal{P}^{j}$ and threshold $T$, while resulting in the increased FLOPs for downstream classification. Generally, the performance of classifier would better for processing more patches, but is sometimes not guaranteed in our empirical study, especially the patches localized by $B^{63}$. We argue that additional patches may introduce some uninformative noises thus damaging the classification decision. Especially the input image that is localized by the branch with narrow RF always includes small object and exists a large proportion of unrelated background, thus is more conspicuous for this observation. Additionally, we provide a reference for tuning the combinations of hyperparameters for trade-off between accuracy and FLOPs, which is quite beneficial for practical deployments and applications.

	\section{Localized Multi-scale Informative Patches}
\begin{figure*}[tbp]
		\centering  
		\includegraphics[width=1.0\linewidth]{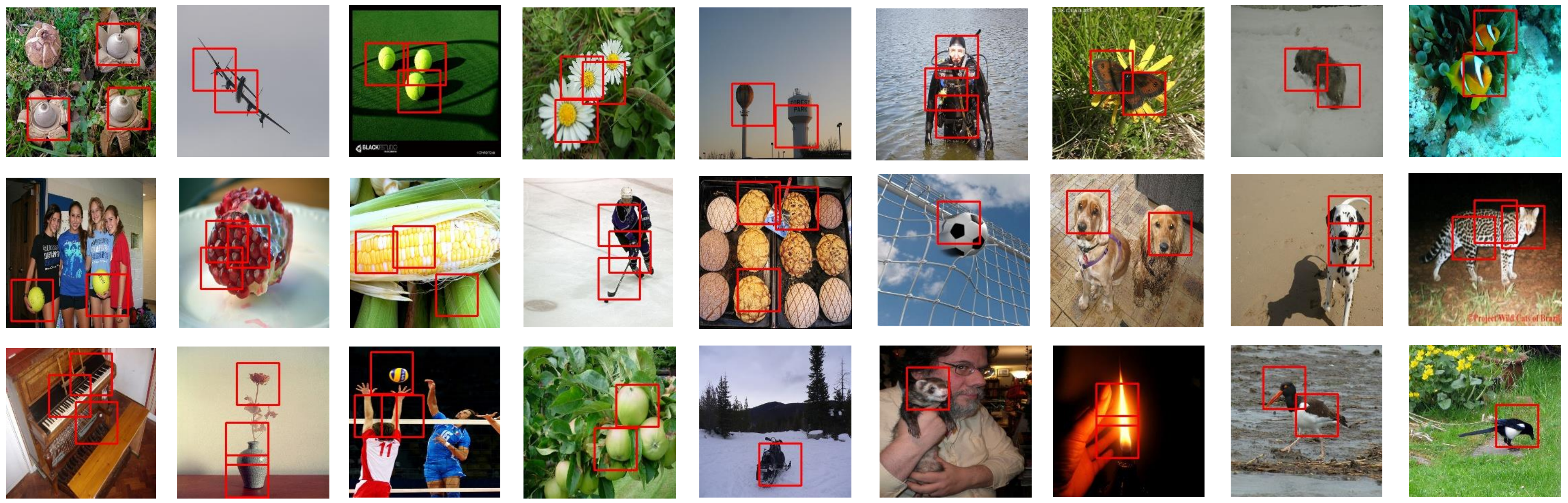}
		\caption{Examples of informative patches with the size of $63\times 63$ localized by branch $B^{63}$.}  
		\label{p_63}
	\end{figure*}

\begin{figure*}[tbp]
		\centering  
		\includegraphics[width=1.0\linewidth]{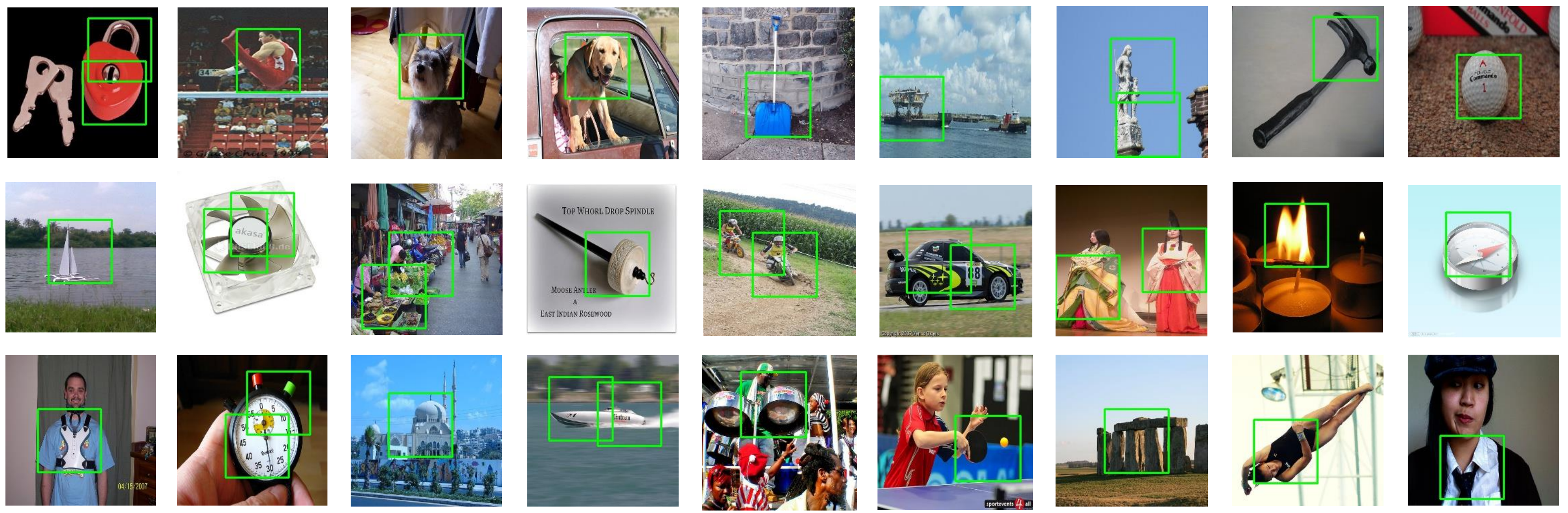}
		\caption{Examples of informative patches with the size of $95\times 95$ localized by branch $B^{95}$.}  
		\label{p_95}
	\end{figure*}

\begin{figure*}[tbp]
		\centering  
		\includegraphics[width=1.0\linewidth]{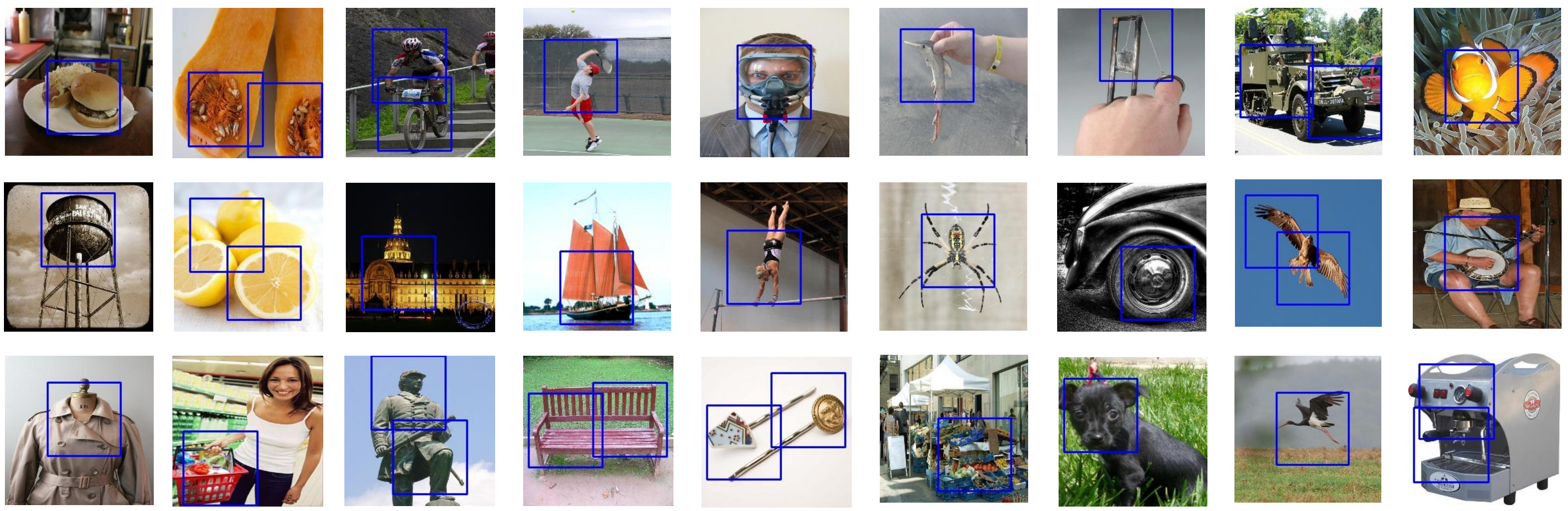}
		\caption{Examples of informative patches with the size of $111\times 111$ localized by branch $B^{111}$.}  
		\label{p_111}
	\end{figure*}

	We display more examples of multi-scale informative image patches localized by the branches of $B^{63}$, $B^{95}$, $B^{111}$ in Figure \ref{p_63}, \ref{p_95}, \ref{p_111}, respectively.
	\begin{figure}[htbp]  
		\centering  
		\includegraphics[width=0.95\linewidth]{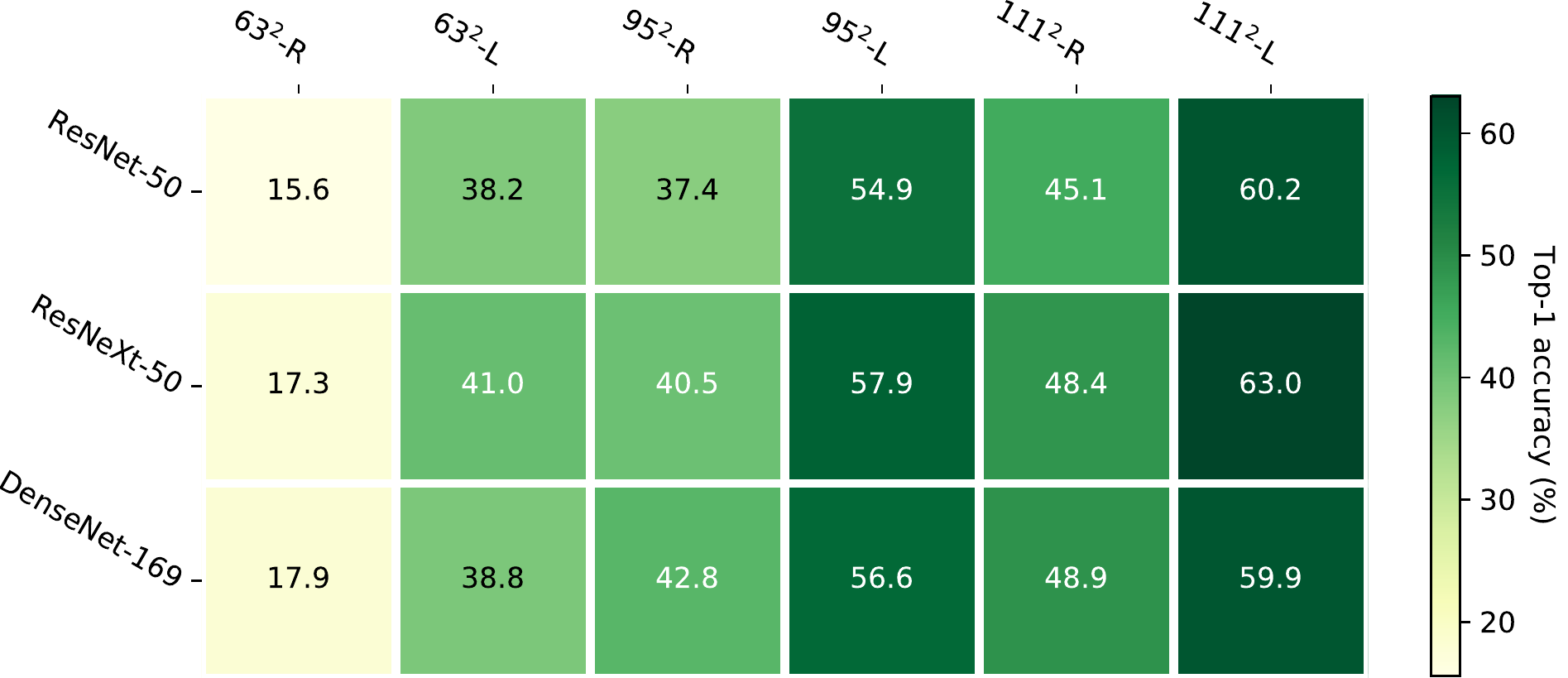}
		\caption{Comparison of top-1 accuracy evaluated on the various downstream CNN models by input patches derived from rescaling (-R) and localizing (-L) across the sizes of $63\times 63$, $95\times 95$ and $111\times 111$.}  
		\label{L_R}
	\end{figure}
	\section{Localized VS Rescaled}
	To further demonstrate the performance of remarkable acceleration and good accuracy is attributed to localizing but unable to be obtained by vanilla scale reduction from the original images, we make a comparison between them and evaluated on the same networks. Each $224\times 224$ image is only localized one informative patch with maximum activation for the correspond scale decided by AnchorNet, meanwhile it is also performed simply rescaling as the counterpart. Figure \ref{L_R} shows that rescaling consistently incurs significant accuracy drop compared with localizing, indicating the localized patches are more informative.
	\section{Robustness to Adversarial Attack}
	\begin{table}
		\caption{Top-1 accuracy of downstream ResNet-50 classifier after FGSM attack.}
		\label{attack}
		\begin{tabular}{lcccc}
			\toprule
			Method&$63\times 63$&$95\times 95$&$111\times 111$& Overall\\
			\midrule
			Baseline & 68.6& 72.1&72.2&71.1\\
			+FGSM & 62.0$_{(\downarrow 6.6)}$& 68.6$_{(\downarrow 3.5)}$& 69.0$_{(\downarrow 3.2)}$& 66.8$_{(\downarrow 4.3)}$\\
			\bottomrule
		\end{tabular}
	\end{table}
	Adversarial machine learning \cite{szegedy2013intriguing, goodfellow2014explaining} has proven that adding imperceptibly small but intentionally worst-case perturbations to the input image changes the prediction of CNN classifier. Goodfellow \emph{et al.}~\shortcite{goodfellow2014explaining} adds the sign of element-wise gradients of loss w.r.t the input along with the perturbation energy $\epsilon$, which is named as Fast Gradient Sigh Method (FGSM). In \shortcite{goodfellow2014explaining}, using FGSM with $\epsilon=0.007$ makes GoogLeNet \cite{szegedy2015going} changing the prediction from panda (correct) to gibbon (incorrect) with high confidence, here $\epsilon$ corresponds to the magnitude of real numbers of the 8 bit image encoding, i.e. [0., 255.]. We further adopt stronger perturbation with $\epsilon=0.2$ to attack AnchorNet so as to experiment the robustness of informative feature localization. The accuracy of downstream classifier ResNet-50 after attacking upstream AnchorNet is shown in Table \ref{attack}. As a reference under this adversarial setting, the classification performance of vanilla ResNet-50 on ImageNet decreases to 44\% from the original 73\%. From Table \ref{attack}, we can observe that the performance of downstream classifier only drops in a relatively small degree compared with vanilla case, which means that the feature localization is quite robust for adversarial attack.
	
\end{document}